%% file: main.tex
\documentclass{article} 
\usepackage{iclr2025_conference,times}

\input{math_commands.tex}

\usepackage[utf8]{inputenc} 
\usepackage[T1]{fontenc}    
\usepackage{hyperref}       
\usepackage{url}            
\usepackage{booktabs}       
\usepackage{amsfonts}       
\usepackage{nicefrac}       
\usepackage{microtype}      
\usepackage{xcolor}         
\usepackage{algorithm}
\usepackage{algpseudocode}
\usepackage{amsmath}
\usepackage{graphicx}
\usepackage{cleveref}
\usepackage{natbib}
\usepackage{subcaption}

\title{DartControl: A Diffusion-Based Autoregressive Motion Model for Real-Time Text-Driven Motion Control}

\author{Kaifeng Zhao, Gen Li, Siyu Tang 
\vspace{2 pt}\\
ETH Zürich\\
\texttt{\{kaifeng.zhao,gen.li,siyu.tang\}@inf.ethz.ch}
}

\newcommand{\update}[1]{\textcolor{black}{#1}}

\iclrfinalcopy 
\begin{document}

\def\methodname{DART}

\maketitle

\input{sections/abstract}

\input{sections/introduction}

\input{sections/related}

\section{Method}
\input{sections/method/preliminary}

\input{sections/method/latent_motion_primitive_model}

\input{sections/method/control}

\input{sections/expr/expr}

\input{sections/conclusion}

\clearpage
\textbf{Acknowledgements.} We sincerely acknowledge the
anonymous reviewers for their insightful feedback. We thank Korrawe Karunratanakul for helpful discussion and suggestions, and Siwei Zhang for meticulous proofreading. This work is supported by the SNSF
project grant 200021 204840 and SDSC PhD fellowship.

\bibliography{iclr2025_conference}
\bibliographystyle{iclr2025_conference}

\newpage
\appendix
\input{sections/appendix}

\end{document}

%% file: math_commands.tex
\usepackage{amsmath,amsfonts,bm}

\def\eqref#1{equation~\ref{#1}}

\def\1{\bm{1}}

\DeclareMathAlphabet{\mathsfit}{\encodingdefault}{\sfdefault}{m}{sl}
\SetMathAlphabet{\mathsfit}{bold}{\encodingdefault}{\sfdefault}{bx}{n}



%% file: sections/abstract.tex
\begin{abstract}
\label{sec:abs}
 
Text-conditioned human motion generation, which allows for user interaction through natural language, has become increasingly popular. Existing methods typically generate short, isolated motions based on a single input sentence. 
However, human motions are continuous and extend over long periods, carrying rich semantics. Creating long, complex motions that precisely respond to streams of text descriptions, particularly in an online and real-time setting, remains a significant challenge.
Furthermore, incorporating spatial constraints into text-conditioned motion generation presents additional challenges, as it requires aligning the motion semantics specified by text descriptions with geometric information, such as goal locations and 3D scene geometry. 
To address these limitations, we propose \textbf{Dart}\textbf{C}ontrol, in short \textbf{DART}, a \textbf{D}iffusion-based \textbf{A}utoregressive motion primitive model for \textbf{R}eal-time \textbf{T}ext-driven motion \textbf{C}ontrol. 
Our model effectively learns a compact motion primitive space jointly conditioned on motion history and text inputs using latent diffusion models.
By autoregressively generating motion primitives based on the preceding motion history and current text input, \methodname{} enables real-time, sequential motion generation driven by natural language descriptions.
%
Additionally,  the learned motion primitive space allows for precise spatial motion control, which we formulate either as a latent noise optimization problem or as a Markov decision process addressed through reinforcement learning. 
We present effective algorithms for both approaches, demonstrating our model’s versatility and superior performance in various motion synthesis tasks. 
Experiments show our method outperforms existing baselines in motion realism, efficiency, and controllability.
Video results and code are available at \href{https://zkf1997.github.io/DART/}{https://zkf1997.github.io/DART/}.
\end{abstract}

%% file: sections/introduction.tex
\section{Introduction}
\label{sec:intro}
Text-conditioned human motion generation has gained increasing popularity in recent years for flexible user interaction via natural languages. Existing text-conditioned motion models ~\citep{tevet2023human, guo2023momask, zhang2023t2m, guoGeneratingDiverseNatural2022, jiang2024motiongpt} primarily focus on generating standalone short motions from a single descriptive sentence.
These methods fail to accurately generate long and complex motions composed of multiple action segments, where each segment is conditioned on distinct action descriptions.
FlowMDM~\citep{barquero2024seamless} is the state-of-the-art temporal motion composition method, capable of generating complex, continuous motions by composing desired actions with precise adherence to their specified durations.
However, FlowMDM is an offline method that requires prior knowledge of the entire action timeline and has a slow generation speed, making it unsuitable for online and real-time applications.
%

In addition to text-based semantic control, generating human motion within spatial constraints and achieving specific goals, such as reaching a keyframe body pose, following a joint trajectory, or interacting with objects, has broad applications but introduces additional complex challenges.
%
Recent works \citep{shafir2024human, karunratanakulOptimizingDiffusionNoise2024, xie2024omnicontrol} have sought to integrate text-conditioned motion models with spatial control capabilities. However, they still face challenges in effectively balancing spatial control, motion quality, and semantic alignment with text. Moreover, these approaches are typically restricted to controlling isolated short motions in an offline setting.
%
Meanwhile, interactive character control~\citep{kovar_motion_2008, holden2015learning,lingCharacterControllersUsing2020d, ASE} has been a longstanding focus in computer graphics, with a primary emphasis on achieving motion realism and responsiveness to interactive control signals.
However, most of these works lack support for text-conditioned semantic control and are limited by being trained on small, curated datasets.
Incorporating text-conditioned motion generation could provide a novel intuitive language interface for animators and everyday users to control the characters, reducing the effort required when specifying detailed spatial control signals is challenging or tedious.



To address these limitations, we propose \textbf{\methodname{}}, a diffusion-based autoregressive motion primitive model for real-time text-driven motion control. 
Moreover, the compact and expressive motion space of \methodname{} provides a foundation for integrating precise spatial control through latent space optimization or reinforcement learning (RL)-based policies.
\methodname{} features three key components. 

First, \methodname{} represents human motion as a collection of motion primitives~\citep{zhang_wanderings_2022}, which are autoregressive representations consisting of overlapping short motion segments tailored for online generation and control. 
These short primitives also provide a more precise alignment with atomic action semantics compared to longer sequences, enabling effective learning of a text-conditioned motion space.
By focusing on shorter primitives, \methodname{} avoids the complexities and extensive data demands of modeling entire motion sequences, allowing for high-quality motion generation with only a few diffusion steps.

Next, \methodname{} learns a text-conditioned autoregressive motion generation model from large-scale data using a latent diffusion architecture, which contains a variational autoencoder for learning a compact latent motion primitive space, and a denoiser network for generating motion primitives conditioned on texts and history.
Leveraging the trained denoiser and decoder models, \methodname~employs an autoregressive rollout to synthesize motion sequences from real-time text inputs, enabling the efficient generation of motions of arbitrary length. Compared with the offline temporal motion composition method FlowMDM, \methodname~provides real-time response and 10x generation speed.

Lastly, we introduce a latent space control framework based on \methodname{} for spatially controllable motion synthesis, leveraging its learned space of realistic human motions to ensure high-quality generation. We present effective optimization and learning algorithms that explore the latent diffusion noise space to synthesize motion sequences that precisely follow both textual and spatial constraints. 
We evaluate \methodname{} across various motion synthesis tasks, including generating long, continuous sequences from sequential text prompts, in-between motion generation, scene-conditioned motion, and goal-reaching synthesis.
The experimental results show that \methodname~is a simple, unified and highly effective motion model, consistently outperforming or matching the performance of baselines. 




%% file: sections/related.tex
\section{Related Works}

\noindent\textbf{Conditional Motion Generation.}
Generating realistic and diverse human motions is a long-standing challenge in computer vision and graphics. 
Apart from generating highly realistic human motions \citep{kovar_motion_2008,holden_learned_2020,buttner_motion_2015,zinno_ml_2019}, conditional generation is another important factor that aligns motion generation with human intentions and various application constraints.
Text-conditioned motion generation \citep{tevet2023human,zhang2022motiondiffuse,petrovichTEMOSGeneratingDiverse2022b,guoGeneratingDiverseNatural2022,jiang2024motiongpt,zhang2023t2m} has become increasingly popular since it allows users to modulate motion generation with flexible natural languages.
Audio and speech-driven motion synthesis methods \citep{alexanderson2023listen, tseng2022edge,siyao2022bailando, ao2022rhythmic, ao2023gesturediffuclip} have also made significant progress recently.
Moreover, there exist many applications that require spatial awareness and precise control in motion generation, such as interactive character control \citep{lingCharacterControllersUsing2020d, ASE, starke_deepphase_2022, luo2023universal, starke2024categorical}, human-scene interactions\citep{hassan_stochastic_2021,starke_neural_2019,zhaoCompositionalHumanSceneInteraction2022b, zhaoSynthesizingDiverseHuman2023a, li2024egogen, xu2023interdiff, jiang2024scaling, wang2024move, liu2023revisit, li2023controllable, zhang2022couch, zhang2024force}, and human-human(noid) interactions \citep{liang2024intergen, zhang2023simulation, christen2023learning, cheng2024expressive, shan2024towards}.
Synthesizing high-quality motions with precise spatial control remains challenging, and \methodname{} is a step toward a general and efficient motion model that supports precise control tasks.


\noindent\textbf{Diffusion Generative Models.}
Denoising Diffusion Models \citep{ho2020denoising, songDenoisingDiffusionImplicit2022, song2020score} are generative models that learn to predict clean data samples by iteratively annealing the noise from a standard Gaussian sample. 
Diffusion models have achieved unprecedented performances in many generation tasks including images, videos, and 3D human motions \citep{tevet2023human, rombachHighResolutionImageSynthesis2022, ho2022video}.
Diffusion models can accept flexible conditions to modulate the generation, such as text prompts, images, audio, and 3D objects \citep{rombachHighResolutionImageSynthesis2022, tevet2023human, alexanderson2023listen,tseng2022edge, zhang2023adding, xu2023interdiff, li2023object}.
Most existing diffusion-based motion generation methods focus on offline generations of short, isolated motion sequences while neglecting the autoregressive nature of human motions \citep{tevet2023human, barquero2024seamless, chen2023executing, karunratanakulOptimizingDiffusionNoise2024, cohan2024flexible, motionlcm, chen2023humanmac}.
Among these methods, DNO~\citep{karunratanakulOptimizingDiffusionNoise2024} is closely related to our optimization-based control approach, as both use diffusion noises as the latent space for editing and control.
However, the key distinction lies in our latent motion primitive-based diffusion model, which, unlike DNO's diffusion model trained with full motion sequences in explicit motion representations, achieves superior performance in harmonizing spatial control with text semantic alignment during experiments.
There are also works that incorporate history conditions in diffusion-based motion generation and capture \citep{xu2023interdiff, jiang2024scaling, van2024diffusionposer, Han2024}.
\citet{shi2024amdm} and \citet{chenTamingDiffusionProbabilistic2024} adapt diffusion models for real-time character motion generation and control. While \citet{shi2024amdm} learns character control policies using diffusion noises as the action space, akin to our reinforcement-learning-based control, their method focuses on single-frame autoregressive generation and lacks support for text conditions, which offers a compact and intuitive interface for users to control character behaviors. 
In contrast, \methodname{} is an efficient and general motion model that scales effectively to large motion-text datasets. \methodname{} supports natural language interfacing and provides a versatile foundation for various motion generation tasks with spatial control.
\update{
Concurrent work, CloSD~\citep{tevet2024closd}, trains autoregressive motion diffusion models conditioned on target joint locations to guide the human motion towards these goals. This approach relies on paired training data of control signals and human motions.
In contrast, \methodname{} learns a latent motion space from motion-only data and introduces latent space control methods to achieve flexible control goals without the need for paired training data.
}

%% file: sections/method/preliminary.tex
\subsection{Preliminaries}

\textbf{Problem Definition.}
We focus on the task of text-conditioned online motion generation with spatial control. Given an $H$ frame seed motion $\mathbf{H}_{seed} = [\mathbf{h}^{1},... ,\mathbf{h}^{H}]$, a sequence of $N$ text prompts $C = [c^{1},... ,c^{N}]$, and spatial goals $g$, the objective is to autoregressively generate continuous and realistic human motion sequences $\mathbf{M}=[\mathbf{H}_{seed}, \mathbf{X}^1, ..., \mathbf{X}^N]$, where each motion segment $\mathbf{X}^i$ matches the semantics of the corresponding text prompt $c^i$ and satisfies the spatial goal constraints~$g$.
This task imposes challenges in high-level action semantic control, precise spatial control, and smooth temporal transition in motion generation.

\textbf{Autoregressive Motion Primitive Representation.} 
We model long-term human motions as the sequential composition of motion primitives \citep{zhang_wanderings_2022} with overlaps for efficient generative learning and online inference. 
Each motion primitive $\mathbf{P}^i=[\mathbf{H}^i, \mathbf{X}^i]$ is a short motion clip containing $H$ frames of history motion $\mathbf{H}^i=[\mathbf{h}^{i, 1},... ,\mathbf{h}^{i, H}]$ that overlap with the previous motion primitive, and $F$ frames of future motion $\mathbf{X}^i=[\mathbf{x}^{i, 1},...,\mathbf{x}^{i, F}]$. 
The history motion of the \textit{i}-th motion primitive $\mathbf{H}^i$ consists of the last $H$ frames of the previous motion primitive $\mathbf{X}^{i-1,F-H+1:F}$.
Therefore, infinitely long motions can be represented as the rollout of such overlapping motion primitives as $\mathbf{M}=[\mathbf{H}_{seed}, \mathbf{X}^1, ..., \mathbf{X}^N]$. 
To represent human bodies in each motion frame, we use an overparameterized representation based on the SMPL-X~\citep{SMPL-X:2019} parametric human body model.
Each frame is represented as a $D=276$ dimensional vector including the body root translation $\mathbf{t}$, root orientation $\mathbf{R}$, local joint rotations $\boldsymbol{\theta}$, joint locations $\mathbf{J}$, and the temporal difference features of locations and rotations.
Each motion primitive is canonicalized in a local coordinate frame centered at the first-frame body pelvis.
We use history length $H=2$ and future length $F=8$ in our experiments.
Further details of the primitive representation are attached in the Appendix \ref{sec:app_primitive}.
For brevity, we omit the primitive index superscript $i$ when discussing in the context of a single primitive.

In contrast to directly modeling long motion sequences, the primitive representation decomposes globally complex sequences into short and simple primitives, resulting in a more tractable data distribution for generative learning. 
Moreover, the autoregressive and simple nature of the primitive representation makes it inherently suitable for fast online generations.
Furthermore, motion primitives convey more interpretable semantics than individual frames, enhancing the learning of text-conditioned motion space.

%% file: sections/method/latent_motion_primitive_model.tex
\subsection{DART: A Diffusion-Based Autoregressive Motion Primitive Model}

\label{sec:lmp}

\begin{figure*}[t]
    \centering
    \includegraphics[width=\linewidth]{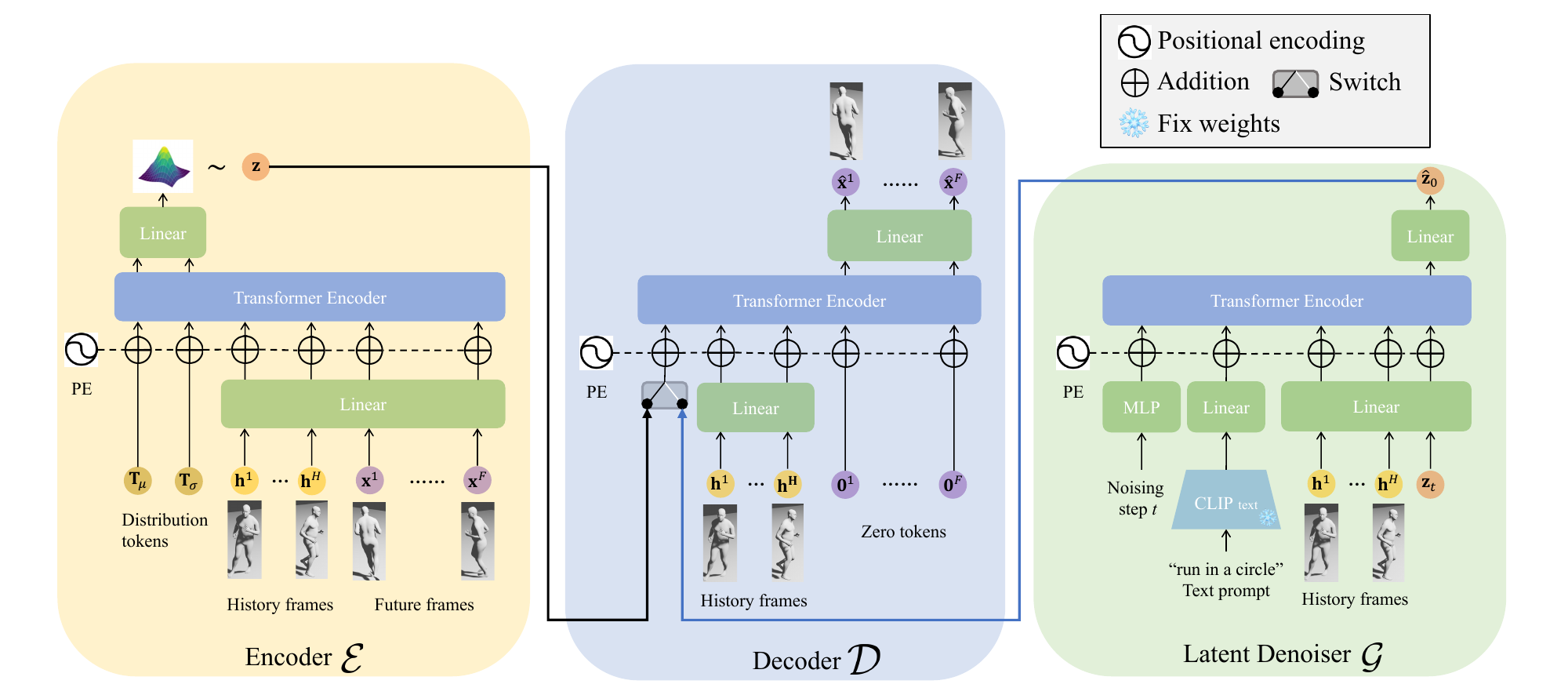}
    \caption{Architecture illustration of \methodname{}. The encoder network compresses the future frames $\mathbf{X}=[\mathbf{x}^{1},...,\mathbf{x}^{F}]$ into a latent variable, conditioned on the history frames $\mathbf{H}=[\mathbf{h}^{1},...,\mathbf{h}^{H}]$. The decoder network reconstructs the future frames conditioned on the history frames and the latent sample. The denoiser network predicts the clean latent sample $\hat{\mathbf{z}}_0$ conditioned on the noising step, text prompt, history frames, and noised latent sample $\mathbf{z}_t$. During the denoiser training, the encoder and decoder network weights remain fixed.} 
\label{fig:model-arch}
\end{figure*}

We propose a latent diffusion model \citep{rombachHighResolutionImageSynthesis2022, chen2023executing} designed for seamless autoregressive motion generation, conditioned on text prompts and motion history. 
The proposed model contains a variational autoencoder (VAE)~\citep{kingmaAutoEncodingVariationalBayes2022a} that compresses the motion primitives into a compact latent space and a latent denoising diffusion model that predicts clean latent variables from noise, conditioned on text prompts and motion history.

\textbf{Learning the Latent Motion Primitive Space.}
We introduce a motion primitive VAE that compresses motion primitives into a compact latent space, upon which we train our latent diffusion models, rather than using the raw motion space \citep{rombachHighResolutionImageSynthesis2022, chen2023executing}. 
The design of learning a compressed latent space of motion primitives is inspired by the observations that raw motion data from the used motion capture dataset AMASS \citep{AMASS:ICCV:2019} often contain \update{various levels of artifacts including glitches and jitters}, and training diffusion models on raw motion space leads to results inheriting such artifacts. This is evidenced by the significantly higher jittering in generated motions of the ablative model without VAE in Appendix.~\ref{sec:app_comose_ablation}.
The compression achieved through the motion primitive VAE significantly mitigates the impacts of \update{these outlier artifacts in motion data}. The resulting latent representation is not only more compact but also more computationally efficient than the raw motion data, thereby enhancing the efficiency of our generative model and improving the control capabilities within the latent space. 

Our motion primitive VAE employs a transformer-based architecture based on \update{MLD}~\citep{chen2023executing}, comprising an encoder $\mathcal{E}$ and decoder $\mathcal{D}$, as shown in Fig.~\ref{fig:model-arch}. 
The encoder takes as input the history motion frames $\mathbf{H}$ and future motion frames $\mathbf{X}$ as well as the learnable distribution tokens $\mathbf{T}_{\mu}$ and $\mathbf{T}_{\sigma}$, which are responsible for predicting the distribution mean and variance.
The latent sample $\mathbf{z}$ is obtained from the predicted distribution via reparameterization \citep{kingmaAutoEncodingVariationalBayes2022a}.
The decoder then predicts the future frames $\mathbf{\hat{X}}$ from zero tokens conditioned on the latent sample $\mathbf{z}$ and the history frames $\mathbf{H}$.
The motion primitive VAE is trained with the future frame reconstruction loss $L_{rec}$, auxiliary losses $L_{aux}$ that penalize unnatural motion reconstruction, 
a small Kullback-Leibler (KL) regularization term $L_{KL}$,
\update{and additional SMPL-based reconstruction and regularization terms}.
We refer to Appendix \ref{sec:app_vae} for further details about the motion primitive VAE.

\textbf{Latent Motion Primitive Diffusion Model.}
Building on the compact latent motion primitive space, we formulate text-conditioned autoregressive motion generation via a probabilistic distribution $q(\mathbf{z} \vert \mathbf{H}, c)$, and train a latent diffusion model $\mathcal{G}$ to approximate it.
Diffusion models \citep{ho2020denoising} are generative models that learn data distributions by progressively reversing a forward diffusion process, which iteratively adds Gaussian noise to data samples until they approach pure noise $\mathcal{N}(\mathbf{0}, \mathbf{I})$.
Given a motion primitive sampled from the dataset and its latent representation $\mathbf{z}_0$ obtained using the encoder $\mathcal{E}$, the forward diffusion produces a sequence of increasingly noisy samples $\mathbf{z}_1, \dots, \mathbf{z}_T$ by iteratively adding noises as:
    $q(\mathbf{z}_t \vert \mathbf{z}_{t-1}) = \mathcal{N}(\sqrt{1-\beta_t} \mathbf{z}_{t-1}, \beta_t \mathbf{I})$, 
where $\beta_t$ are noise schedule hyper parameters.
The denoiser model learns the reverse process 
$p_{\theta}(\mathbf{z}_{t-1} \vert \mathbf{z}_t, t, \mathbf{H}, c)=\mathcal{N}(\boldsymbol{\mu}_t, \boldsymbol{\Sigma}_t)$ 
for generating motion primitives conditioned on the motion history $\mathbf{H}$ and text label $c$ of the motion primitive.
The vairance $\boldsymbol{\Sigma}_t$ of the reverse process distribution is scheduled using hyper parameters, while the mean $\boldsymbol{\mu}_t$ is modeled using a denoiser neural network. We design the denoiser model $\mathcal{G}$ to predict the clean latent variable $\hat{\mathbf{z}}_0=\mathcal{G}(\mathbf{z}_t, t, \mathbf{H}, c)$, from which  
$\boldsymbol{\mu}_t$ can be derived as follows: 
\begin{equation}
\boldsymbol{\mu}_t = \frac{\sqrt{\bar{\alpha}_{t-1}}\beta_t}{1 - \bar{\alpha}_t} \mathcal{G}(\mathbf{z}_t, t, \mathbf{H}, c) + \frac{\sqrt{\alpha_t}(1 - \bar{\alpha}_{t-1})}{1 - \bar{\alpha}_t} \mathbf{z}_t,
\end{equation}
where $\alpha_t = 1 - \beta_t$ and $\bar{\alpha}_t = \prod_{i=1}^t \alpha_i$.
During generation, we initialize with $\mathbf{z}_T \sim \mathcal{N}(\mathbf{0}, \mathbf{I})$ and use the denoiser model to predict the clean variable $\hat{\mathbf{z}}_0$, which is subsequently diffused to a lower noise level $\mathbf{z}_{T-1}$.This denoising process iterates until a clean sample $\mathbf{z}_0$ is obtained.

The denoiser model architecture is shown in Fig.~\ref{fig:model-arch}. 
%
The diffusion step $t$ is embedded using a small MLP, while the text prompt $c$ is encoded using the CLIP \citep{radfordLearningTransferableVisual2021a} text encoder. The text prompt is randomly masked out by a probability of 0.1 during training to enable classifier-free guidance \citep{hoClassifierFreeDiffusionGuidance2022} during generation. 
The cleaned latent variable $\hat{\mathbf{z}}_0$ can be converted back to the future frames using the frozen decoder $\mathcal{D}$: $\hat{\mathbf{X}}=\mathcal{D}(\mathbf{H},\hat{\mathbf{z}}_0)$.
We train the latent denoiser $\mathcal{G}$ using the simple objective~\citep{ho2020denoising}.
Additionally, we apply the reconstruction loss $L_{rec}$ and auxiliary losses $L_{aux}$ on the decoded future frames $\hat{\mathbf{X}}$. 
Notably, we use only 10 diffusion steps for both training and inference. This small number suffices for realistic sample generation due to the simplicity of our motion primitive representation, enabling highly efficient online generation.
Moreover, we use the scheduled training \citep{lingCharacterControllersUsing2020d, bengioScheduledSamplingSequence2015, rempe2021humor} to progressively introduce the test-time distribution of the history motion $\mathbf{H}$, which improves the stability of long sequence online generation and the text prompt controllability for unseen poses.
We refer to Appendix~\ref{sec:app_denoiser} for the details.
With the trained motion primitive decoder $\mathcal{D}$, latent denoiser $\mathcal{G}$ and a diffusion sampler $\mathcal{S}$ such as \update{DDPM and DDIM} \citep{ho2020denoising, songDenoisingDiffusionImplicit2022}, we can autoregressively generate motion sequences 
given the history motion seed $\mathbf{H}_{seed}$ and the online sequence of text prompts $C$, as shown in Alg.~\ref{alg:rollout}.
During sampling, we use classifier-free guidance \citep{hoClassifierFreeDiffusionGuidance2022} on the text condition with a guidance scale $w$:
$
    \mathcal{G}_w(\mathbf{z}_t, t, \mathbf{H}, c) = \mathcal{G}(\mathbf{z}_t, t, \mathbf{H}, \emptyset) + w \cdot (\mathcal{G}(\mathbf{z}_t, t, \mathbf{H}, c) - \mathcal{G}(\mathbf{z}_t, t, \mathbf{H}, \emptyset)).
$
Using the rollout algorithm, \methodname{} generates over 300 frames per second using a single RTX 4090 GPU,  enabling real-time applications and online reinforcement-learning control as in Sec.~\ref{sec:control}.

\begin{algorithm}[t]
\caption{Autoregressive rollout generation using latent motion primitive model}
\begin{algorithmic}
\algrenewcommand\algorithmiccomment[1]{\hfill\(\triangleright\) #1}
    \State \textbf{Input:} primitive decoder $\mathcal{D}$, latent variable denoiser $\mathcal{G}$, history motion seed $\mathbf{H}_{seed}$, text prompts $C=[c^{1},...,c^{N}]$, total diffusion steps $T$, classifier-free guidance scale $w$, diffusion sampler $\mathcal{S}$.
    \State \textbf{Optional Input:} Latent noises $\mathbf{Z}_T=[\mathbf{z}_T^1,...,\mathbf{z}_T^N]$
    \State \textbf{Output:} motion sequence $\mathbf{M}$
    
    \State $\mathbf{H} \gets \mathbf{H}_{seed}$
    \State $\mathbf{M} \gets \mathbf{H}_{seed}$
    \For{$i \gets 1$ \textbf{to} $N$} \Comment{number of rollouts}
        \State sample noise $\mathbf{z}_T^i$ from $\mathcal{N}(0, 1)$ if not inputted
        \State $\hat{\mathbf{z}}_0^i \gets \mathcal{S}(\mathcal{G}, \mathbf{z}_T^i, T,  \mathbf{H}, c^i, w$)  \Comment{diffusion sample loop with classifier-free guidance}
        \State $\hat{\mathbf{X}} \gets \mathcal{D}(\mathbf{H}, \hat{\mathbf{z}}_0^i)$ 
        \State $\mathbf{M} \gets \Call{concat}{\mathbf{M}, \hat{\mathbf{X}}}$  \Comment{concatenate future frames to generated sequence}
        \State $\mathbf{H} \gets \Call{canonicalize}{\hat{\mathbf{X}}^{F-H+1:F}}$ \Comment{update the rollout history with last H generated frames}
    \EndFor
    \State return $\mathbf{M}$
\end{algorithmic}
\label{alg:rollout}
\end{algorithm}

%% file: sections/method/control.tex
\subsection{Spatially Controllable Motion Synthesis via DART}

\label{sec:control}

Text-conditioned motion generation offers a user-friendly interface for controlling motions through natural language.
However, relying solely on text limits precise spatial control, such as walking to a specific location or sitting in a designated spot.
Therefore, it is necessary to incorporate motion control mechanisms to achieve precise spatial goals, including reaching a keyframe body pose, following joint trajectories, and interacting with scene objects.
We formulate the motion control task as generating the motion sequence $\mathbf{M}$ that minimizes its distance to a given spatial goal $g$ under a criterion function $\mathcal{F}(\cdot, \cdot)$ and the regularization from the scene and physical constraints $cons(\cdot)$:

\begin{equation}
    \mathbf{M}^* = {\arg\!\min}_{\mathbf{M}} \mathcal{F}(\Pi(\mathbf{M}), g) + cons(\mathbf{M}),
    \label{equ:motion control}
\end{equation}

where $g$ is the task-dependent spatial goal (e.g., a keyframe body for motion in-between tasks or a target location for navigation tasks),
$\Pi(\cdot)$ is the projection function that extracts goal-relevant features from motion sequences and maps them into the task-aligned observation space,
and $cons(\cdot)$ denotes physical and scene constraints, such as preventing scene collisions and floating bodies.

Directly solving the motion control task in the raw motion space often results in unrealistic motions since most samples in the raw motion space do not represent plausible motions. To improve the generated motion quality, many previous methods \update{tackle} such motion control tasks in a latent motion space, where samples can be mapped to plausible motions \citep{karunratanakulOptimizingDiffusionNoise2024,lingCharacterControllersUsing2020d, ASE, holden2015learning}. 
\methodname{} offers a powerful text-conditioned latent motion space for such latent space control, as it learns a generative model capable of producing diverse and realistic motions from standard Gaussian samples.
Using the deterministic DDIM \citep{songDenoisingDiffusionImplicit2022} sampler, we adapt \methodname{} sampling to function as a deterministic mapping from latent noises $\mathbf{Z}_T$ to plausible motions.
This allows us to reformulate the motion control task in Eq.~\ref{equ:motion control} as a latent space control problem as follows: 
Given the initial motion history $\mathbf{H}_{seed}$, a sequence of text prompts $C$, the pretrained \methodname{} models, and a deterministic diffusion sampler $\mathcal{S}$, the rollout function in Alg.~\ref{alg:rollout} can deterministically map a list of motion primitive latent noises $\mathbf{Z}_T = [\mathbf{z}_T^1, .., \mathbf{z}_T^N]$ to a motion sequence conditioned on the history motion seed and text prompts:
$
    \mathbf{M} = \Call{rollout}{\mathbf{Z}_T, \mathbf{H}_{seed}, C}.
$
The minimization objective is converted as:
\begin{equation}
    {\mathbf{Z}_T}^* = {\arg\!\min}_{\mathbf{Z}_T} \mathcal{F}(\Pi(\Call{rollout}{\mathbf{Z}_T, \mathbf{H}_{seed}, C}), g) + cons(\Call{rollout}{\mathbf{Z}_T, \mathbf{H}_{seed}, C})
    \label{equ:latent control}
\end{equation}

Note that we do not use DDIM to skip diffusion steps at sampling, which we observe to cause artifacts in generated motion. 
We then propose two solutions to this latent space motion control problem, one is to directly optimize the latent noises using gradient descent, and the other is to model the control task as a Markov process and use reinforcement learning to learn control policies.

\textbf{Motion Control via Latent Diffusion Noise Optimization.}
One straightforward solution to this minimization problem (Eq.~\ref{equ:latent control}) is to directly optimize the latent noises $\mathbf{Z}_T$ given the criterion function using gradient descent methods~\update{\citep{kingmaAdamMethodStochastic2017, karunratanakul2023dno}}. 
The latent noise optimization is illustrated in Alg.~\ref{alg:latetn_opt}.
This optimization-based control framework is general and applicable for various spatial control tasks. 
We instantiate the latent noise optimization method in two example control scenarios: in-between motion generation and human-scene interaction generation.
\begin{algorithm}[t]
\caption{Latent noises optimization}
\begin{algorithmic}
\algrenewcommand\algorithmiccomment[1]{\hfill\(\triangleright\) #1}
    \State \textbf{Input:} Latent noises $\mathbf{Z}_T=[\mathbf{z}_T^1,...,\mathbf{z}_T^N]$, Optimizer $\mathcal{O}$, learning rate $\eta$, and goal $g$. 
    (For brevity, we do not reiterate the inputs of the rollout function defined in Alg.~\ref{alg:rollout} and criterion terms in Eq.~\ref{equ:motion control})
    \State \textbf{Output:} a motion sequence $\mathbf{M}.$
    \For{$i \gets 1$ \textbf{to} optimization steps} 
        
        \State $\mathbf{M} \gets \Call{rollout}{\mathbf{Z}_T,  \mathbf{H}_{seed}, C}$  
        \State $\nabla \gets \nabla_{\mathbf{Z}_T} (\mathcal{F}(\Pi(\mathbf{M}), g) + cons(\mathbf{M}))$
        \State $ \mathbf{Z}_T \gets \mathcal{O}(\mathbf{Z}_T, \nabla / \|\nabla\|, \eta)$ \Comment{update using normalized gradient}
    \EndFor
    \State return $\mathbf{M} \gets \Call{rollout}{\mathbf{Z}_T,  \mathbf{H}_{seed}, C}$
\end{algorithmic}
\label{alg:latetn_opt}
\end{algorithm}

First, we address the motion in-between task that aims to generate the motion frames transition between given history and goal keyframes $g$ that is $f$ frames away conditioned on the text prompt~$c$. 
We use the distance between the $f$-th frame of the generated motion and the goal keyframe as the optimization objective.
Second, we show that physical and scene constraints $cons(\cdot)$ can be incorporated to synthesize human motions in a contextual environment. Given \update{input 3D scenes}, text prompts $C$, and spatial goals $g$ of the interaction anchor joint locations, e.g.,~locations of the pelvis when sitting, the objective is to generate motions that not only perform the desired interaction but also achieve the goal joint positions while adhering to scene constraints. During the optimization, \update{the 3D scenes are represented as signed distance fields (SDF) to compute body-scene distances, which serve as the basis for deriving human-scene contact and collision metrics that encourage foot-floor contact and scene collision avoidance}, as detailed in Appendix.~\ref{sec:app_optim}.





\textbf{Motion Control via Reinforcement Learning.}
Although the proposed latent noise optimization is effective for general control tasks, the optimization can be computationally expensive.
To address this, the autoregressive primitive-based motion representation of \methodname{} allows for another efficient control mechanism using reinforcement learning (RL) \citep{sutton_introduction_1998}. We model the latent motion control as a Markov decision process with a latent action space and use RL to learn policy models to achieve the goals.
We model digital humans as agents to interact with an environment according to a policy to maximize the expected discounted return. At each time step $i$, the agent observes the state $\mathbf{s}^i$ of the system, samples an action $\mathbf{a}^i$ from the learned policy, with the system transitioning to the next state $\mathbf{s}^{i+1}$ due to the performed action $\mathbf{a}^i$, and receives a reward $r^i = r(\mathbf{s}^i, \mathbf{a}^i, \mathbf{s}^{i+1})$. 

\begin{figure}[t]
    \centering
    \includegraphics[width=0.8\linewidth]{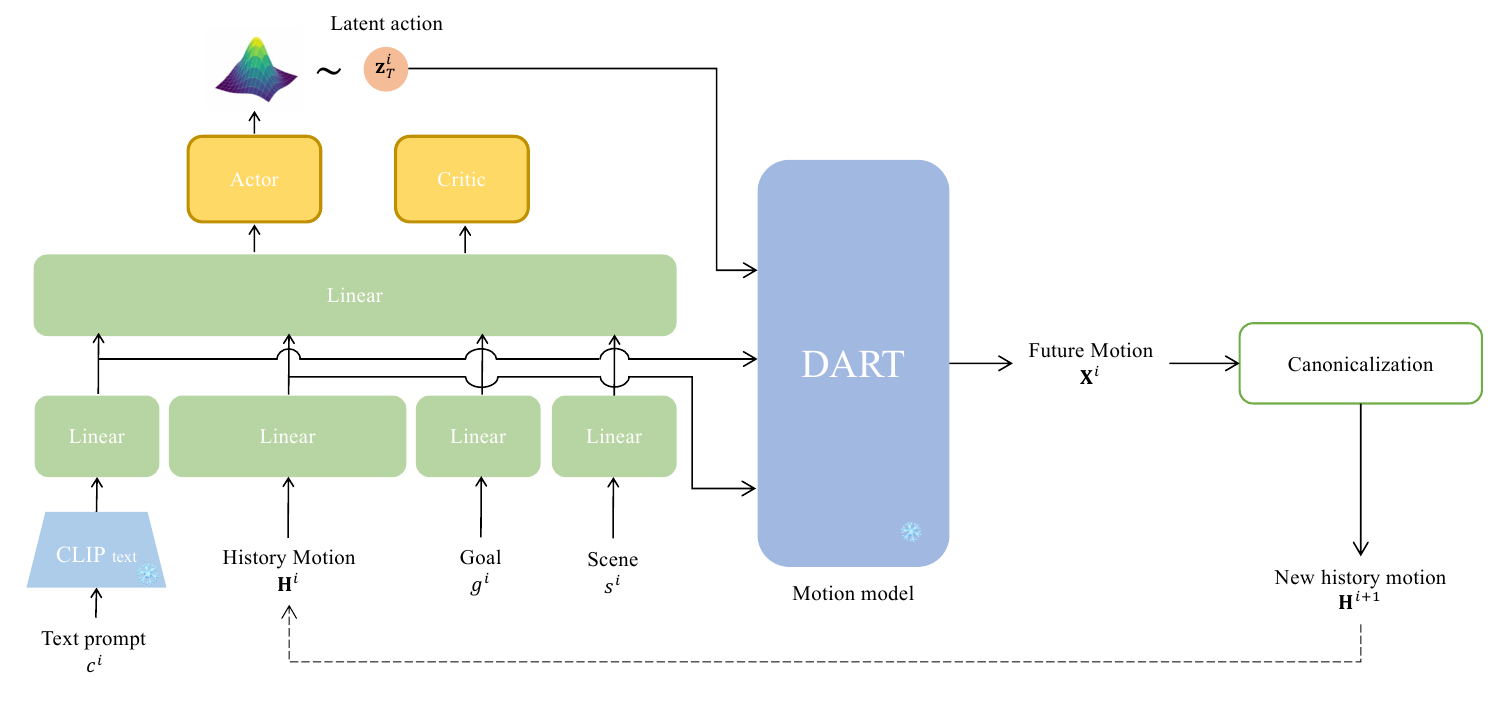}
    \caption{Architecture of the reinforcement learning-based control policy. The pretrained \methodname{} diffusion denoiser and decoder models transform the latent actions into motion frames. The last predicted frames are canonicalized and provided to the policy model as the next step history condition.} 
\label{fig:rl}
\end{figure}
Our latent motion primitive model naturally fits in the Markov decision process due to its autoregressive nature.
We use the latent noises $\mathbf{z}_T$ as the latent action space and train goal-conditioned policy models as controllers.
The policy architecture is shown in Fig.~\ref{fig:rl}.
Our policy model employs the actor-critic \citep{sutton_introduction_1998} architecture and is trained with the PPO \citep{schulman2017proximal} algorithm.
The state $\mathbf{s}^i$ includes the history motion observation $\mathbf{H}^i$, goal observation $g^i$, scene observation $s^i$, and the CLIP embedding of the text prompt $c^i$. The policy model takes in [$\mathbf{H}^i$, $g^i$, $s^i$, $c^i$] to predict the latent noise $\mathbf{z}_T^i$ as the action. The latent noise $\mathbf{z}_T^i$ is mapped to the future motion frames $\mathbf{X}^i$ using the frozen latent denoiser $\mathcal{G}$ and motion primitive decoder $\mathcal{D}$. The new history motion is extracted from the last $H$ predicted frames and fed to the policy network in the next step.
We reformulate the minimization problem in Eq.~\ref{equ:latent control} as reward maximization to train the policy.

We instantiate the reinforcement learning control with the text-conditioned goal-reaching task.
Given a text prompt $c$ and a 2D goal location $g$, we aim to control the human to reach the goal location using the action specified by the text.
The goal location is transformed into a local observation, which includes its distance to the body pelvis at the last history frame and its local direction within the human-centric coordinate frame. We consider a simple flat scene and the scene observation is the relative floor height to the body pelvis at the first history frame.
The policy is trained with distance rewards encouraging the human pelvis to reach the goal location and scene constraint rewards penalizing foot skating and floor penetration. Further details can be found in \cref{sec:app_rl}.
With the trained control policies, we can efficiently control a human to reach dynamic goals using specified skills like walking or hopping.

%% file: sections/expr/expr.tex
\section{Experiments}
We provide extensive experiments showing how \methodname{} can serve as a general model for text-conditioned temporal motion composition (\ref{sec:exp_compose}) and various motion generation tasks requiring precise spatial control via latent noise optimization (\ref{sec:exp_opt}) and reinforcement learning policy (\ref{sec:exp_rl}). 
Qualitative results and comparisons are available in the supplementary videos on the \href{https://zkf1997.github.io/DART/}{project page}.

Our \methodname{} is trained on motion-text data from the BABEL~\citep{punnakkalBABELBodiesAction2021c} dataset in our experiments if not otherwise stated. BABEL contains motion capture sequences with frame-aligned text labels that annotate the fine-grained semantics of actions. 
Fine-grained text labels in BABEL allow models to learn precise human action controls and natural transitions among actions.
However, \methodname{} can also learn using motions with coarse sequence-level labels such as the HML3D~\citep{guoGeneratingDiverseNatural2022} dataset, as in the optimization-based motion in-between experiments in Sec.~\ref{sec:inbetween}.

\input{sections/expr/composition}
\input{sections/expr/optimization}
\input{sections/expr/RL}

%% file: sections/expr/composition.tex
\subsection{Text-Conditioned Temporal Motion Composition}
\label{sec:exp_compose}
Text-conditioned temporal motion composition aims to generate realistic motion sequences that faithfully align with a list of action segments, each defined by a specific text prompt and duration.
We evaluate the motion composition task on the BABEL\citep{punnakkalBABELBodiesAction2021c} dataset consisting of motion capture sequences with human-annotated per-frame action descriptions, which facilitate the evaluation of precise action controls and natural transitions in motion composition.
Since BABEL does not release the test set, we compare our \methodname{} with baseline methods on the BABEL validation set. We extract the list of action segments described by tuples of text prompts and durations from each data sequence and feed the action lists as conditions for motion composition. 
 
We evaluate the generation results using metrics proposed in \citet{guoGeneratingDiverseNatural2022} and \citet{barquero2024seamless}.
For each action segment, we evaluate the similarity between generation and dataset (FID), motion-text semantic alignment (R-prec, MM-DIST), and generation diversity (DIV).
To evaluate smooth transitions between two segments, we measure the jerk (the derivative of acceleration) of the 30-frame transition clip centered at the splitting point of two action segments, reporting the peak jerk (PJ) and Area Under the Jerk (AUJ). 
Moreover, we profile all methods in a benchmark of generating one 5000-frame-long sequence and report the generation speed, the latency of getting the first generated frames, and memory usage.
We also conduct human preference studies to evaluate motion realism and motion-text semantic alignment of generation results, during which the participants are given generation results from two different methods and are asked to select the generation that is perceptually more realistic or better aligns with the action text stream in subtitles.
We compare \methodname{} with baselines inlcuding TEACH \citep{athanasiouTEACHTemporalAction2022}, DoubleTake~\citep{shafir2024human}, a history-conditioned modification of T2M-GPT~\citep{zhang2023t2m}(denoted as T2M-GPT*), and the state-of-the-art offline motion composition method FlowMDM~\citep{barquero2024seamless}.

We present the quantitative results in Tab.~\ref{tab:online} and Tab.~\ref{tab:amt_study}.
\methodname{} achieves the best FID in both the segment and transition evaluation, indicating the highest similarity to the dataset and best motion realism. 
\methodname{} also displays second-best jerk metrics indicating smooth action transitions.
We observe that \methodname{} performs slightly worse than FlowMDM in motion-text semantic alignment (R-prec and MM-Dist) because of the online generation nature of \methodname{}. Natural action transitions require time to transit to the new action after receiving the new action prompt, leading to a delay in the emergence of the new action semantics. For instance, a human cannot immediately transition from kicking a leg in the air to stepping backward without first recovering to a standing pose.
This inherent delay in transitions results in a motion embedding shift that impacts the R-prec metric of \methodname{}.
In contrast, the offline baseline FlowMDM generates the entire sequence as a whole and requires oracle information of the full timeline of action segments to modulate compatibility between subsequent segments.
The slight but natural action transition delay of \methodname{} is perceived as natural by humans, as shown in the preference study results in Tab.~\ref{tab:amt_study}. \methodname{} is preferred over all the baselines, including FlowMDM, for both motion realism and motion-text semantic alignment in human evaluations.




\methodname{} requires significantly less memory than the offline baseline FlowMDM and achieves approximately 10x faster generation, with a frame rate exceeding 300 and a latency of 0.02s, enabling real-time text-conditioned motion composition (see supplementary video).
We refer to Appendix~\ref{sec:app_compose} for details of experiments, user studies, and ablation studies about model architecture and hyperparameters, and refer to the supplementary video for qualitative comparisons.

\begin{table}[ht]
\centering
\tiny
\setlength\tabcolsep{2pt}
\caption{Quantitative evaluation results on text-conditioned temporal motion composition. 
The first row includes the metrics of the dataset for reference.
Symbol `$\rightarrow$' denotes that closer to the dataset reference is better and `$\pm$'  indicates the 95\% confidence interval.
\textbf{Bold} and {\color{blue}{blue}} texts indicate the best and second best results excluding the dataset, respectively.
}




\begin{tabular}{l|cccc|cccc|ccc}
\hline
 & \multicolumn{4}{c|}{Segment} & \multicolumn{4}{c|}{Transition} & \multicolumn{3}{c}{Profiling} \\
 & FID$\downarrow$ & R-prec$\uparrow$ & DIV $\rightarrow$ & MM-Dist$\downarrow$ & FID$\downarrow$ & DIV $\rightarrow$ & PJ$\rightarrow$ & AUJ $\downarrow$ & Speed(frame/s)$\uparrow$& Latency(s)$\downarrow$ & Mem.(MiB)$\downarrow$ \\
\hline
Dataset & 0.00$\pm$0.00 & 0.72$\pm$0.00 & 8.42$\pm$0.15 & 3.36$\pm$0.00 & 0.00$\pm$0.00 & 6.20$\pm$0.06 & 0.02$\pm$0.00 & 0.00$\pm$0.00 &  & \\
\hline
TEACH  & 17.58$\pm$0.04 & {\color{blue}{0.66}}$\pm$0.00 & 10.02$\pm$0.06 & 5.86$\pm$0.00 & 3.89$\pm$0.05 & 5.44$\pm$0.07 & 1.39$\pm$0.01 & 5.86$\pm$0.02 & \textbf{3880}$\pm$144&{\color{blue}{0.05}}$\pm$0.00 & 2251 \\
DoubleTake & 7.92$\pm$0.13 & 0.60$\pm$0.01 & \textbf{8.29}$\pm$0.16 & 5.59$\pm$0.01 & 3.56$\pm$0.05 & \textbf{6.08}$\pm$0.06 & 0.32$\pm$0.00 & 1.23$\pm$0.01 & 85$\pm$1 & 59.11$\pm$0.76 & \textbf{1474} \\
T2M-GPT* & 7.71$\pm$0.55 & 0.49$\pm$0.01 & 8.89$\pm$0.21 & 6.69$\pm$0.08 & 2.53$\pm$0.04 & {\color{blue}{6.61}}$\pm$0.02 & 1.44$\pm$0.03 & 4.10$\pm$0.09 & {\color{blue}885}$\pm$12 &0.23 $\pm$0.00  & {\color{blue}2172} \\
FlowMDM & {\color{blue}{5.81}}$\pm$0.10 & \textbf{0.67}$\pm$0.00 & 8.90$\pm$0.06 & \textbf{5.08}$\pm$0.02 & {\color{blue}2.39}$\pm$0.01 & 6.63$\pm$0.08 & \textbf{0.04}$\pm$0.00 & \textbf{0.11}$\pm$0.00 & 31$\pm$0 & 161.29$\pm$0.24 & 11892  \\
\hline
Ours & \textbf{3.79}$\pm$0.06 & 0.62$\pm$0.01 & {\color{blue}{8.05}}$\pm$0.10 & {\color{blue}{5.27}}$\pm$0.01 & \textbf{1.86}$\pm$0.05 & 6.70$\pm$0.03 & {\color{blue}{0.06}}$\pm$0.00 & {\color{blue}{0.21}}$\pm$0.00 & 334 $\pm$2 & \textbf{0.02}$\pm$0.00 & 2394 \\
\hline
\end{tabular}

\label{tab:online}
\end{table}

\begin{table}[ht]
\centering
\caption{Human preference study results comparing our method against baselines in generation realism and motion-text semantic alignment on text-conditioned temporal motion composition. We report the percentage of each method being voted better than the other (Ours vs. Baselines). }
\scriptsize
\begin{tabular}{lcc}
\toprule
 & Realism (\%) & Semantic (\%) \\
\midrule
Ours vs. TEACH & \textbf{66.7} vs. 33.3 & \textbf{66.0} vs. 34.0 \\
Ours vs. DoubleTake & \textbf{66.4} vs. 33.6 & \textbf{66.1} vs. 33.9 \\
Ours vs. T2M-GPT* & \textbf{61.3} vs. 38.7 & \textbf{66.7} vs. 33.3 \\
Ours vs. FlowMDM & \textbf{53.3} vs. 46.7 & \textbf{51.3} vs. 48.7 \\
\bottomrule
\end{tabular}
\label{tab:amt_study}
\end{table}

%% file: sections/expr/optimization.tex
\subsection{Latent Diffusion Noise Optimization-Based Control Using DART}
\label{sec:exp_opt}


\paragraph{Text-conditioned motion in-between.}
\label{sec:inbetween}
Motion in-betweening aims to generate realistic motion frames that smoothly transition between a pair of history and goal keyframes. We consider a text-conditioned variant where an additional text prompt is inputted to specify the action semantics of the frames in between.
We compare our method with DNO \citep{karunratanakulOptimizingDiffusionNoise2024} and OmniControl \citep{xie2024omnicontrol}.
For a fair comparison, we train \methodname{} on the HML3D dataset same as the baselines. 
We evaluate using test sequences covering diverse actions, with the sequence lengths ranging from 2 to 4 seconds. 
The quantitative evaluations are shown in Tab.~\ref{table:in-between}. 
We report the $L_2$ norm errors between the generated motion and the history motion and goal keyframe. We also evaluate the motion realism with the skate and jerk metrics. The skate metric \citep{lingCharacterControllersUsing2020d,zhangModeadaptiveNeuralNetworks2018} calculates a scaled foot skating when in contact with the floor: $s=disp \cdot (2-2^{h/0.03})$, where $disp$ is the foot displacement in two consecutive frames, $h$ denotes the higher foot height in consecutive frames and 0.03m is the threshold value for contact.
We do not calculate skate metric for sequences where the feet are not on a flat floor, such as crawling and climbing down stairs.
Our method can generate the motions closest to the keyframe and show fewer skating and jerk artifacts. 
Our method effectively preserves the semantics specified by the text prompts, while the baseline DNO occasionally ignores the text prompts to reach the goal keyframe, as illustrated in the examples of pacing in circles and dancing in the supplementary video.
This highlights the superior capability of our latent motion primitive-based \methodname{} in harmonizing spatial control and text semantic alignment.



\begin{table}[h!]
\scriptsize
\centering
\caption{Quantitative evaluation of text-conditioned motion in-between. The best results excluding the dataset are in \textbf{bold} and `$\pm$'  indicates the 95\% confidence interval.
}


\begin{tabular}{l|c|c|c|c}
\hline
 & History error (cm)\textbf{$\downarrow$} & Goal error (cm)\textbf{$\downarrow$} & Skate (cm/s)$\downarrow$ &Jerk\textbf{$\downarrow$} \\ \hline
Dataset & 0.00 $\pm$ 0.00 & 0.00 $\pm$ 0.00 & 2.27 $\pm$ 0.00 & 0.74 $\pm$ 0.00 \\
\hline
OmniControl & 21.22 $\pm$ 2.86 & 7.79 $\pm$ 1.91 & 4.97 $\pm$ 1.31 & 1.41 $\pm$ 0.08 \\ 
DNO & 1.20 $\pm$ 0.20 & 4.24 $\pm$ 1.34 & 5.38 $\pm$ 0.70 & 0.65 $\pm$ 0.06 \\ 
Ours & \textbf{0.00} $\pm$ 0.00 & \textbf{0.59} $\pm$ 0.01 & \textbf{2.98} $\pm$ 0.32 & \textbf{0.61} $\pm$ 0.01\\ 
\hline
\end{tabular}

\label{table:in-between}
\end{table}

\begin{figure}[ht]
\centering
\begin{minipage}{\textwidth}
    \centering
    \begin{subfigure}[b]{0.3\textwidth}
        \centering
        \includegraphics[width=\textwidth]{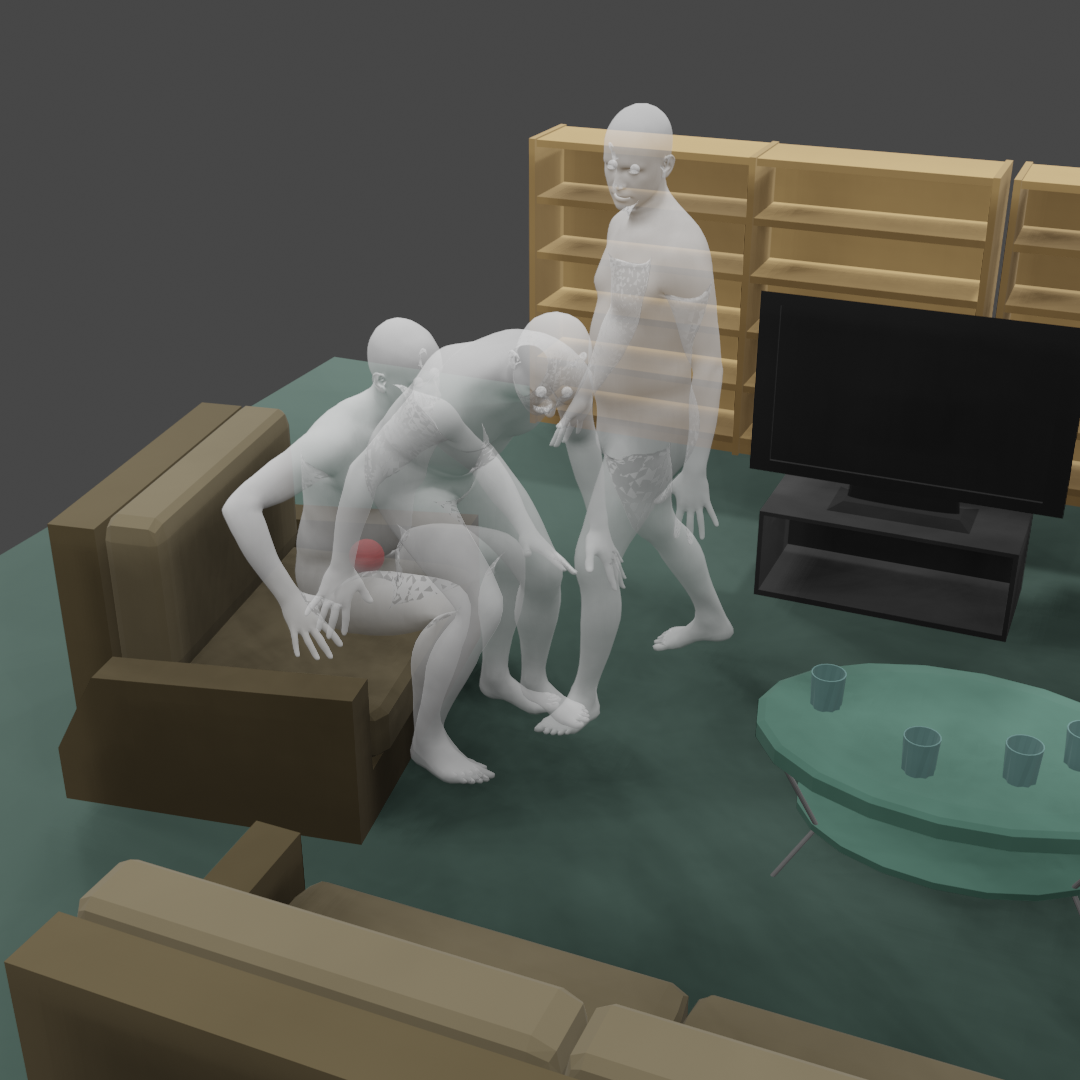}
        \caption{Walk, turn left, sit on the chair}
    \end{subfigure}
    \begin{subfigure}[b]{0.3\textwidth}
        \centering
        \includegraphics[width=\textwidth]{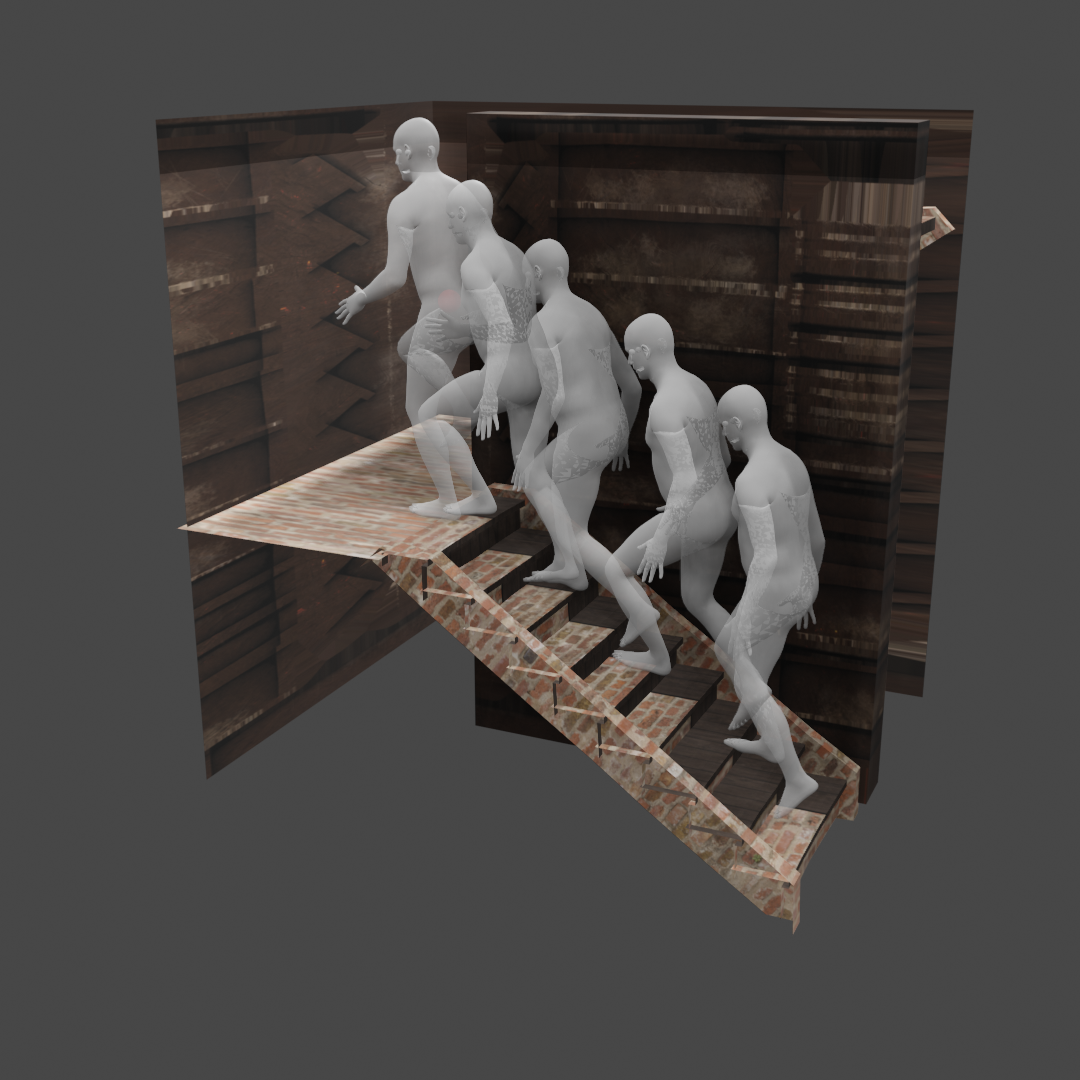}
        \caption{Walk upstaris}
    \end{subfigure}
    \begin{subfigure}[b]{0.3\textwidth}
        \centering
        \includegraphics[width=\textwidth]{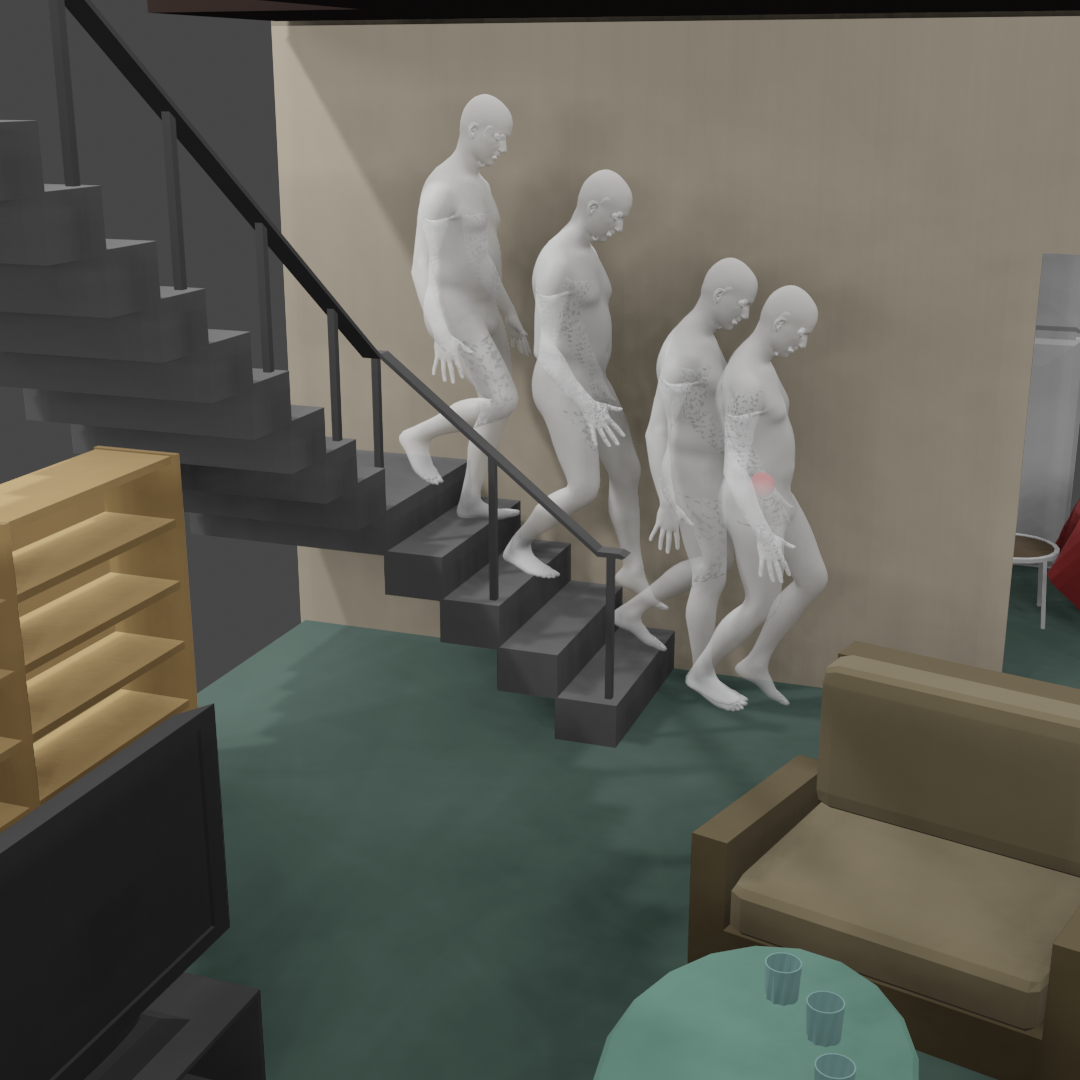}
        \caption{Walk downstaris}
    \end{subfigure}
    \caption{Illustrations of human-scene interaction generation given text prompts and goal pelvis joint location (visualized as a red sphere). Best viewed in the supplementary video.}
    \label{fig:interaction}
\end{minipage}
\end{figure}

\paragraph{Human-scene interaction.}
We qualitatively show that our latent noise optimization control can be applied to human-scene interaction synthesis, where the goal is to control the human to interact naturally with the surrounding environment. Given \update{an input 3D scene and} the text prompts specifying the actions and durations, we use latent noise optimization to control the human to reach the goal joint location while adhering to the scene contact and collision constraints.
\update{The input scenes are represented as signed distance fields for evaluating human-scene collision and contact constraints as detailed in Appendix~\ref{sec:app_optim}}.
We present generated interactions of climbing stairs and walking to sit on a chair in Fig.~\ref{fig:interaction} and the supplementary video.

%% file: sections/expr/RL.tex
\subsection{Reinforcent Learning-Based Control Using DART}
\label{sec:exp_rl}
By integrating \methodname{} with reinforcement learning-based control, we train text-conditioned goal-reaching policy models capable of three locomotion styles: `walk', `run', and `hop on the left leg'.
We evaluate goal-reaching on paths consisting of sequences of waypoints.
We compare our method to a baseline GAMMA \citep{zhang_wanderings_2022}, which
also trains goal-reaching policies with a learned motion action space. Unlike \methodname{}, GAMMA lacks text conditioning and is limited to generating walking motions.
The evaluation metrics include the reach time, the success rate of reaching the final goal waypoint, foot skating, and foot-floor distance.
The evaluation results are shown in Tab.~\ref{table: goal_reach}.
Our policy consistently reaches all goals within a reasonable timeframe, while GAMMA occasionally fails to meet the final goal and may float off the floor beyond the contact threshold.
Moreover, our text-conditioned goal-reaching policy achieves a generation speed of \textbf{240 frames per second}.
These results demonstrate the potential of \methodname{} as a foundational human motion model, upon which versatile control models for various tasks can be learned through reinforcement learning. 

\begin{table}[h!]
\scriptsize
\centering
\caption{Quantitative evaluation of text-conditioned goal-reaching controller. The best results are in \textbf{bold} and `$\pm$'  indicates the 95\% confidence interval.}
\begin{tabular}{l|c|c|c|c}
\hline
 & Time (s)$\downarrow$ & Success rate$\uparrow$ & Skate (cm/s) $\downarrow$ & Floor distance (cm)$\downarrow$ \\ 
 \hline
GAMMA walk & $31.44 \pm 2.58$ & $0.95 \pm 0.03$ & $5.14 \pm 1.58$ & $5.55 \pm 0.84$ \\ 

\textcolor{black}{Ours~`walk'} & \textcolor{black}{$17.08 \pm 0.05$} & \textcolor{black}{$\mathbf{1.0} \pm 0.0$} & \textcolor{black}{$2.67 \pm 0.12$} & \textcolor{black}{$\textbf{2.24} \pm 0.02$} \\ 
\hline
\textcolor{black}{Ours~`run'} & \textcolor{black}{$\mathbf{10.55} \pm 0.06$} & \textcolor{black}{$\mathbf{1.0} \pm 0.0$} & \textcolor{black}{$3.23 \pm 0.24$} & \textcolor{black}{$3.86 \pm 0.05$} \\ 
\textcolor{black}{Ours~`hop on left leg'} & \textcolor{black}{$20.50 \pm 0.24$} & \textcolor{black}{$\mathbf{1.0} \pm 0.0$} & \textcolor{black}{$\textbf{2.22} \pm 0.12$} & \textcolor{black}{$4.11 \pm 0.07$} \\

\hline
\end{tabular}
\label{table: goal_reach}
\end{table}

%% file: sections/conclusion.tex
\section{Limitations and conclusions}
\label{sec:conclusion}

\methodname{} relies on motion sequences with frame-level aligned text annotations, as in BABEL, to achieve precise text-motion alignment and natural transition between actions.
When trained on the coarse sentence-level motion labels from HML3D, the text-motion alignment degenerates for texts describing multiple actions, resulting in motions randomly switching between the described actions in a random order. This occurs because each short motion primitive inherently matches only a portion of the sequence's semantics. Using a coarse sentence-level description as the text label for a primitive causes semantic misalignment and ambiguity. We aim to explore hierarchical latent spaces to effectively tackle both fine-grained and global sequence-level semantics~\citep{stoffl2024behavemae} in the future. 

Our method \methodname~effectively learns a text-conditioned motion primitive space that enables real-time online motion generation driven by natural languages. 
Additionally,  the learned powerful motion primitive space allows for precise spatial motion control via latent noise optimization or reinforcement learning policies. 
Experiments demonstrate the superiority of \methodname{} in harmonizing spatial control with motion text-semantic alignment in the generated motions.

%% file: sections/appendix.tex
\appendix

\section{Motion Primitive Representation}
\label{sec:app_primitive}
\textbf{Representation.}
We represent each frame of the motion primitive as a tuple of $(\mathbf{t}, \mathbf{R}, \boldsymbol{\theta}, \mathbf{J}, \mathrm{d}\mathbf{t}, \mathrm{d}\mathbf{R}, \mathrm{d}\mathbf{J})$, where $\mathbf{t} \in \mathbb{R}^3$ denotes the global body translation, $\mathbf{R} \in \mathbb{R}^6$ denotes the 6D rotation representation \citep{zhouContinuityRotationRepresentations2020} of the global body orientation, $\boldsymbol{\theta} \in \mathbb{R}^{21 \times 6}$ is the 6D representation of 21 joint rotations, $\mathbf{J} \in \mathbb{R}^{22 \times 3} $ denotes the 22 joints locations, $\mathrm{d}\mathbf{t} \in \mathbb{R}^3$ denotes the temporal difference with previous frame's translation, $\mathrm{d}\mathbf{R} \in \mathbb{R}^6$ denotes the 6D representation of the relative rotation between current and previous frame's body orientation, and $\mathrm{d}\mathbf{J} \in \mathbb{R}^{22 \times 3} $ denotes the temporal diffence between current and previous frame's joint locations.

Our motion representation is \update{overparameterized \citep{holden2017phase, lingCharacterControllersUsing2020d, rempe2021humor, guoGeneratingDiverseNatural2022}}, with multiple benefits. Firstly, the joint rotation components $\boldsymbol{\theta}$ are compatible with animation pipelines, saving the time-consuming optimization-based skeleton-to-body conversion required by the commonly used HML3D \citep{guoGeneratingDiverseNatural2022} representation in text-to-motion methods.
Moreover, including the joint location components $\mathbf{J}$ facilitates solving physical constraints like reducing foot skating and joint trajectory control. Our motion representation also models the first-order kinematics with the temporal difference features to improve the motion naturalness.

\textbf{Canonicalization.}
We represent motion primitives in a human-centric local coordinates frame to canonicalize the primitive features and facilitate model learning.
Each motion primitive is canonicalized in a local coordinates system centered at the first frame body.
The origin is located at the pelvis of the first frame body, the X-axis is the horizontal projection of the vector pointing from the left hip to the right hip, and the Z-axis is pointing in the inverse gravity direction.
Given the pelvis, left hip, and right hip joints, the origin is located at the pelvis joint, and the local axis system can be derived by:

\begin{algorithm}[h]
\caption{Motion primitive rotation canonicalization}
\begin{algorithmic}
\algrenewcommand\algorithmiccomment[1]{\hfill\(\triangleright\) #1}
    \State \textbf{Input:} right\_hip $\in \mathbb{R}^3$, left\_hip $\in \mathbb{R}^3$
    \State x\_axis = right\_hip - left\_hip
    \State x\_axis[2] = 0 \Comment{Project to the xy plane}
    \State normalize(x\_axis)
    \State z\_axis = [0, 0, 1] \Comment{Inverse gravity direction}
    \State y\_axis = cross\_porudct(z\_axis, x\_axis)
    \State normalize(y\_axis)
    \State return [x\_axis, y\_axis, z\_axis]
\end{algorithmic}
\label{alg:canonical}
\end{algorithm}

We store the canonicalization transformations and apply their inverse transformation to the generated motion primitives to recover global motions in the world coordinates.

\section{Datasets}
We train separate \methodname{} models on motion-text data from the BABEL~\citep{punnakkalBABELBodiesAction2021c} and HML3D~\citep{guoGeneratingDiverseNatural2022} dataset. Both BABEL and HML3D use motion sources from the AMASS~\citep{AMASS:ICCV:2019} dataset. Their main difference is that BABEL features fine-grained frame-level text annotation while HML3D uses coarse sequence-level annotation.

The BABEL dataset contains motion capture sequences with frame-aligned text labels that annotate the fine-grained semantics of actions. 
Fine-grained text labels in BABEL allow models to learn precise human action controls and natural transitions among actions.
We use the motion data at a framerate of 30 frames per second the same as prior works~\citep{barquero2024seamless, athanasiouTEACHTemporalAction2022}.
During training, motion primitives are randomly sampled from data sequences and the text label is randomly sampled from all the action segments that overlap with the primitive.
To alleviate the action imbalance in the BABEL dataset, we use the provided action labels to perform importance sampling during training so that the motion data of each action category has roughly equal sampling chances despite their varying frequency in the original dataset.
To maintain compatibility with prior works, we fix the human gender to male and set the body shape parameter to zero.
 
The HML3D dataset contains short motions with coarse sequence-level sentence descriptions. We only use a subset of HML3D in training since its subset HumanAct12 and left-right mirroring motion sequences only provide joint locations instead of SMPL~\citep{SMPL:2015, SMPL-X:2019} body sequences that our motion representation requires. The joint rotations used in the original HML3D are calculated using naive inverse kinematics and can not be directly used to animate human motions. We therefore only train using the subset where the SMPL body motion sequences is available. 
The HML3D motions are sampled at a framerate of 20 frames per second. During training, we randomly sample primitives with uniform probability and the text label is randomly chosen from one of the multiple sentence captions of the overlapping sequence.
To maintain compatibility with prior works, we fix the human gender to male and set the body shape parameter to zero.

\section{Motion Primitive VAE}
\label{sec:app_vae}
\textbf{Architecture.}
Our Motion Primitive VAE employs the transformer-based architecture. Both the encoder and decoder consist of 7 transformer encoder layers with skip connections \citep{chen2023executing}.
The transformer layers use the dropout rate of 0.1, feed-forward dimension of 1024, hidden dimension of 256, 4 attention heads, and the gelu activation function.
The latent space dimension is 256. After finishing training, we follow \citep{rombachHighResolutionImageSynthesis2022} to calculate the variance of the latent variables using a data batch. When training the latent denoiser model, the raw latent output from the encoder is scaled to have a unit standard deviation.

\textbf{Losses.}
The  motion primitive VAE is trained with the following losses:
\begin{equation}
    L_{VAE} = L_{rec} + w_{KL} \times L_{KL} + w_{aux} \times L_{aux} 
    \update{+ w_{SMPL} \times L_{SMPL}}.
\end{equation}

The reconstruction loss $L_{rec}$ aims to minimize the distance between the reconstructed future frames $\hat{\mathbf{X}}$ and the ground truth future frames $\mathbf{X}$ as follows:
\begin{equation}
    L_{rec} = \mathcal{F}(\hat{\mathbf{X}}, \mathbf{X}),
\end{equation}
where $\mathcal{F}(\cdot, \cdot)$ denotes the distance function and we use the smoothed L1 loss\citep{girshickFastRCNN2015} in implementation.

The Kullback-Leibler divergence (KL) term $L_{KL}$ penalizes the distribution difference between the predicted distribution and a standard Gaussian:
\begin{equation}
\label{equ:kl}
\mathcal{L}_{KL} = KL(q(\mathbf{z}\vert \mathbf{H}) \parallel \mathcal{N}(0, \mathbf{I})),
\end{equation}
where $KL(\cdot, \cdot)$ denotes the Kullback-Leibler divergence (KL), and $q(\mathbf{z} \vert \mathbf{H})$ denotes the predicted distribution from the encoder $\mathcal{E}$.
We use a small KL term of $1e^{-6}$ following \citet{rombachHighResolutionImageSynthesis2022} as we aim to keep the latent space expressive and only use the small KL loss to avoid arbitrarily high-variance latent spaces.

The auxiliary loss $L_{aux}$ regularizes the predicted temporal difference features $\hat{\mathrm{d}\cdot}$ of translations, global orientation, and joints to be close to the actual temporal differences $\bar{\mathrm{d} \cdot}$ calculated from the predicted motion features:
\begin{equation}
    L_{aux} = \mathcal{F}(\bar{\mathrm{d}\mathbf{t}}, \hat{\mathrm{d}\mathbf{t}}) + \mathcal{F}(\bar{\mathrm{d}\mathbf{J}}, \hat{\mathrm{d}\mathbf{J}}) + \mathcal{F}(\bar{\mathrm{d}\mathbf{R}}, \hat{\mathrm{d}\mathbf{R}}).
\end{equation}
\update{For instance, the difference of the first two frames of the predicted root translation ($\bar{\mathrm{d}\mathbf{t}}^0:=\hat{\mathbf{t}}^1 - \hat{\mathbf{t}}^0$) should be consistent with the predicted first frame root translation difference feature $\hat{\mathrm{d}\mathbf{t}}^0$.}
We use $w_{aux}=100$ in our experiments. 

\update{The SMPL losses $L_{SMPL}$ consist of two components: the SMPL-based joint reconstruction loss $L_{joint\_rec}$, and the joint consistency loss $L_{consistency}$. The SMPL losses $L_{SMPL}$ are defined as follows:}
\begin{equation}
    \update{L_{SMPL} = L_{joint\_rec} + L_{consistency}.}
    \label{equ:loss_smpl}
\end{equation}
\update{The SMPL joint reconstruction loss $L_{joint\_rec}$ penalizes the discrepancy between the joints regressed from the predicted body parameters and those regressed from the ground truth body parameters:}
\begin{equation}
    \update{L_{joint\_rec} = \mathcal{F}(\mathcal{J}(\hat{\mathbf{t}}, \hat{\mathbf{R}}, \hat{\boldsymbol{\theta}}), \mathcal{J}(\mathbf{t}, \mathbf{R}, \boldsymbol{\theta})),}
\end{equation}
\update{where $\mathcal{J}$ denotes the SMPL body joint regressor that predicts joint locations given the body parameters. The SMPL body shape parameter $\beta$ is assumed to be zero and ignored for simplicity here.}

\update{The joint consistency loss $L_{consistency}$ regularizes the predicted joint locations and predicted SMPL body parameters consistently represent the same body joints:}
\begin{equation}
    \update{L_{consistency} = \mathcal{F}(\hat{\mathbf{J}}, \mathcal{J}(\hat{\mathbf{t}}, \hat{\mathbf{R}}, \hat{\boldsymbol{\theta}})).}
\end{equation}
\update{We use $w_{SMPL}=100$ in our experiments. }
We train the motion primitive VAE with the AdamW \citep{kingmaAdamMethodStochastic2017} optimizer and the learning rate is set to $1e^{-4}$ with linear annealing.

We conduct ablation studies on the impacts of the losses. We evaluate motion primitive VAEs trained with different loss weights on autoregressive motion sequence reconstruction error and motion jittering in reconstructed motions. Using large KL loss weight of $w_{KL}=1$, reduces the model expressiveness (the reconstruction error increases from 0.08 to 0.44 compared to $w_{KL}=1e^{-6}$ on the test motions) and fails to accurately reconstruct complex sequences like cartwheeling. Using a small KL loss weight of $w_{KL}=1e^{-6}$ can maintain the expressiveness of the learned VAE while allowing the latent distribution to deviate a bit more from a standard Gaussian. Applying the auxiliary losses with $w_{aux}=100$ helps to reduce the jittering in the reconstructions and improve the motion quality compared to using $w_{aux}=0$, as reflected by a smaller jerk metric of 2.45 when using $w_{aux}=100$ compared to a jerk of 3.67 when using $w_{aux}=0$.

\section{Latent Denoiser Model}
\label{sec:app_denoiser}
\subsection{Losses}
We train the latent denoiser model using DDPM~\citep{ho2020denoising} with 10 diffusion steps and use a cosine noise scheduler. The denoiser model is trained with the following losses:
\begin{equation}
    L_{denoiser} = L_{simple} + w_{rec} \times L_{rec} + w_{aux} \times L_{aux},
\end{equation}

\begin{equation}
    L_{simple} =  \mathbb{E}_{(\mathbf{z}_0, c) \sim q(\mathbf{z}_0, c), \, t \sim [1, T], \, \epsilon \sim \mathcal{N}(\mathbf{0}, \mathbf{I})}
  \mathcal{F}(\mathcal{G}(\mathbf{z}_t, t, \mathbf{H}, c), \mathbf{z}_0), 
\end{equation}

where $\mathcal{F}(\cdot, \cdot)$ is a distance function and we use the smooth L1 loss \citep{girshickFastRCNN2015} in our implementation.
We train the denoiser to predict the clean latent variable with the simple objective $L_{simple}$, and apply the feature reconstruction loss $L_{rec}$ and auxiliary losses $L_{aux}$ on the decoded motion primitive  $\hat{\mathbf{X}} = \mathcal{D}(\mathcal{G}(\mathbf{z}_t, t, \mathbf{H}, c))$ to ensure the decoded motion primitives are valid.

\subsection{Scheduled Training}
\label{sec:app_schedule}
We use scheduled training to improve the stability of long sequence generation and the text prompt controllability.
Long-term prediction stability is a significant challenge in autoregressive generation since the sample distribution can drift and accumulate during autoregressive generation. When the sample drifts out of the distribution covered by the learned model, the generation results can go wild. Our latent motion primitive model also faces the long-term stability challenge as an autoregressive method.
%

To alleviate the out-of-distribution problems, we use the scheduled training \citep{lingCharacterControllersUsing2020d,bengioScheduledSamplingSequence2015,rempe2021humor, martinez2017human} to progressively introduce the test-time distributions during training.
Specifically, we train the latent denoiser model on sequences of $N$ consecutive motion primitives and use the prediction result of the previous motion primitive instead of the ground truth dataset to extract the history motion input $\mathbf{H}$.
We name such history motion extracted from the predicted last primitive as rollout history.
Using the rollout history instead of the ground truth history introduces the test-time distribution which can differ from the dataset distribution, e.g., unseen human poses or out-of-distribution combinations of human bodies and text labels. Exposing the model to such test-time distribution at training can improve the long-term generation stability and increase the text controllability when facing novel combinations of history motion and text prompts at generation time.

The scheduled training has three stages to progressively introduce the rollout history. The first stage is fully supervised training where only the ground truth history is used during training.
The second scheduled learning stage randomly replaces the ground truth history motion with rollout history motion by a probability linearly increasing from 0 to 1. The third stage of rollout training always uses the rollout history instead of the ground truth history.
The scheduled training algorithm for the latent denoising model is shown in Alg.~\ref{alg:schedule}.

\begin{algorithm}[ht]
\caption{Calculate rollout probability}
\begin{algorithmic}[1]
\State \textbf{Input:} current iteration number $iter$, number of train iterations in the first supervised stage  $I_1$, number of train iterations in the second scheduled stage $I_2$
\State \textbf{Output:} rollout probability $p$
\Function{rollout\_probability}{$iter$, $I_1$, $I_2$}
\If{$iter \leq I_1$}  \Comment{no rollout in the first supervised stage}
    \State $p \gets 0$
    \ElsIf {$iter > I_1 + I_2$} \Comment{the third rollout stage always use rollout}
        \State $p \gets 1$
    \Else  \Comment{linearly scheduled rollout probability in the second scheduled stage}
        \State $p \gets \frac{iter - I_1}{I_2}$
    \EndIf
\State \Return $p$
\EndFunction
\end{algorithmic}
\label{alg:prob}
\end{algorithm}

\begin{algorithm}[ht]
\caption{Scheduled training for the latent denoising model}
\begin{algorithmic}[1]
\State \textbf{Input:} pretrained motion primitive decoder $\mathcal{D}$ and encoder $\mathcal{E}$, latent variable denoiser $\mathcal{G}_{\theta}$ parameterized by ${\theta}$, total diffusion steps $T$, optimizer $\mathcal{O}$, loss criterion $\mathcal{L}$, train dataset $\mathcal{X}$.
\State \textbf{Scheduled training parameters:} the number of train iterations in the first supervised stage  $I_1$, the number of train iterations in the second scheduled stage $I_2$, the number of train iterations in the third rollout stage $I_3$, the maximum number of primitive rollouts $N$ during training.
\State
\State $I_{total} \gets I_1 + I_2 + I_3$
\State $iter \gets 0$
\While{$iter < I_{total}$}
    \State $[\mathbf{H}_{seed}, \mathbf{X}^1, c^{1},...,\mathbf{X}^N, c^{N}] \sim \mathcal{X}$
    \State \Comment{sample $N$ consecutive motion primitives with text labels from dataset $\mathcal{X}$ } 
    \State $\mathbf{H} \gets \mathbf{H}_{seed}$  \Comment{initialize motion history}
    \For{$i \gets 1$ \textbf{to} $N$} \Comment{number of rollouts}
        \State $\mathbf{z}_0^i = \mathcal{E}(\mathbf{H}, \mathbf{X}^i)$  \Comment{compress motion primitive into latent space}
        \State $t \sim \mathcal{U}[0, T)$ \Comment{sample diffusion step $t$}
        \State $\mathbf{z}_t^i \gets \Call{forward\_diffusion}{\mathbf{z}_0^i, t}$
        \State $\hat{\mathbf{z}}_0^i = \mathcal{G}_{\theta}(\mathbf{z}_t^i, t, \mathbf{H}, c^i$)  \Comment{latent denoising model prediction}
        \State $\hat{\mathbf{X}} = \mathcal{D}(\mathbf{H}, \hat{\mathbf{z}}_0^i)$ \Comment{decode predicted latent variable to future motion frames} 
        \State $\nabla \gets \nabla_{\theta}  \mathcal{L}(\mathbf{z}_0^i, \hat{\mathbf{z}}_0^i, \mathbf{H}, \mathbf{X}^i,\hat{\mathbf{X}}^i)$  \Comment{model parameter gradient calculation}
        \State $ \theta \gets \mathcal{O}(\theta, \nabla)$  \Comment{model update using optimizer}
        \State 
        
        \State $p \gets$ \Call{rollout\_probability}{$iter, I_1, I_2$}  
        \State \Comment{update history motion using predicted or GT motion by a scheduled probability}
        \If{$\text{rand()} < p$} \Comment{use predicted rollout history}
            \State $\mathbf{z}_T^i \gets \Call{forward\_diffusion}{\mathbf{z}_0^i, T}$  \Comment{maximum noising simulating inference time} 
            \State $\hat{\mathbf{z}}_0^i = \Call{ddpm\_sample}{\mathcal{G}_{\Theta}, \mathbf{z}_T^i, T,  \mathbf{H}, c^i}$  \Comment{full DDPM denoising loop}
            \State $\hat{\mathbf{X}} = \Call{sg}{\mathcal{D}(\mathbf{H}, \hat{\mathbf{z}}_0^i)}$ \Comment{decode predicted latent variable and stop gradient}
            \State $\mathbf{H} \gets \Call{canonicalize}{\hat{\mathbf{X}}^{F-H+1:F}}$  
        \Else \Comment{use ground truth history}
            \State $\mathbf{H} \gets \Call{canonicalize}{\mathbf{X}^{F-H+1:F}}$ 
        \EndIf
        \State
        \State $iter \gets iter + 1$
    \EndFor
\EndWhile
\end{algorithmic}
\label{alg:schedule}
\end{algorithm}

We use $w_{rec}=1$ and $w_{aux}=10000$ in our experiments.
We do not use the SMPL losses $L_{SMPL}$ in training the denoiser model because the SMPL body model inference process slows down the training. 
The denoiser model is trained using an AdamW optimizer. The learning rate is set to $1e^{-4}$ with linear annealing.
Our denoiser model is trained with scheduled training \citep{lingCharacterControllersUsing2020d, bengioScheduledSamplingSequence2015}, consisting of a fully supervised stage of 100K iterations, a scheduled stage of 100K iterations, and a rollout stage of 100K iterations.
We set the maximum number of rollouts as 4.
With the scheduled training, our latent motion primitive model can stably generate long motion sequences and better respond to the text prompt control even at poses that are not paired with the text prompt in the dataset.

\section{Text-conditioned temporal motion composition}
\label{sec:app_compose}
\subsection{Experiment details}
We apply \methodname{} to conduct online motion generation using the rollout algorithm \ref{alg:rollout} with a default seed motion $\mathbf{H}_{seed}$ of rest standing and using a classifier-free guidance \update{weight of 5}.
We use the released checkpoints of FlowMDM~\citep{barquero2024seamless}, TEACH~\citep{athanasiouTEACHTemporalAction2022}, and DoubleTake~\citep{shafir2024human} for baseline comparison.
We adjust the handshake size and blending length of DoubleTake to be compatible with the shortest segment length of 15 frames.
The history-conditioned modification of T2M-GPT is retrained on the BABEL dataset using the original hyperparameters. At generation time, the last frames of the previous action segment are encoded into tokens and provided as the first tokens when generating the next action segment to provide history conditioning of the previous action.

We extract the timeline of action segments for evaluation from the BABEL~\citep{punnakkalBABELBodiesAction2021c} valid set.
For each sequence in the validation set, we sort the original frame labels provided by BABEL to obtain a list of tuples of text prompts and durations. We randomly sample one text label when a segment is annotated by multiple texts. We skip the `transition' text labels due to the ambiguous semantics and clamp the duration length with a minimum of 15 frames.

\subsection{Human Preference Studies.}
\begin{figure}[ht]
\centering
    \centering
    \includegraphics[width=0.75\linewidth]{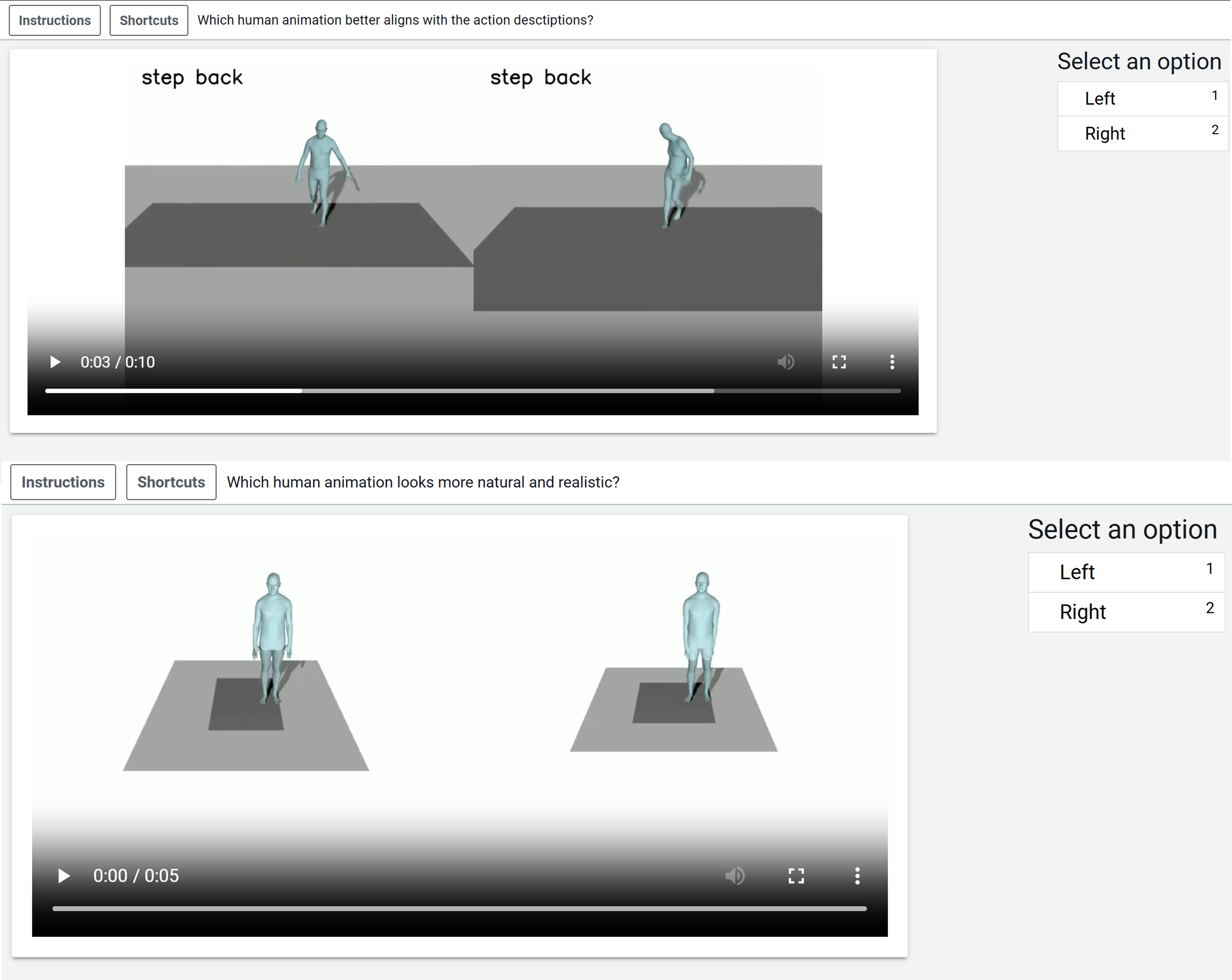}
    \caption{Illustration of the human preference study interface for evaluating motion-text semantic alignment (top) and perceptual realism(bottom). Participants are requested to select the generation that is perceptually more realistic or better aligns with the action descriptions in subtitles (only visible in semantic preference study).}
    \label{fig:amt}
\end{figure}

We conduct human preference studies to quantitatively compare our method DART with baselines to provide a more comprehensive and convincing evaluation. We run human preference studies on Amazon Mechanical Turk(AMT) to evaluate generation realism and text-motion semantic alignment. Participants are given generation results from two different methods and are asked to select the generation that is perceptually more realistic or better aligns with the action text stream in subtitles, as illustrated in Fig.~\ref{fig:amt}. During the realism evaluation, action descriptions were not displayed to eliminate distractions. 
We sample 256 sequences of action texts and durations from the BABEL dataset and use each method to generate motions given the action timelines. Participants are shown random pairs of results from our method and a baseline, and they are asked to choose the better generation. Each comparison is voted by 3 independent participants, with the video pairs randomly shuffled to ensure that participants do not know the method source. 

\subsection{Ablation Studies.}
\label{sec:app_comose_ablation}
We conduct ablative studies on architecture designs, diffusion steps, primitive representation, and scheduled training on the text-conditioned motion composition task.

Removing the variational autoencoder (\methodname{}-VAE) and training diffusion model in the raw motion space results in a significantly higher jittering as reflected by the peak jerk (PJ) and Area Under Jerk (AUJ) metrics. This validates the effectiveness of integrating the variational encoder to compress the high-frequency noise in motion data and improve motion generation quality.

Without the scheduled training (\methodname{}-schedule), the model can not effectively respond to text control and have significantly worse R-Prec and FID metrics. This is because, without scheduled training, the model can easily encounter out-of-distribution combinations of history motion and text prompts during autoregressive generation.

We also include the ablation study of training a model to predict the single next frame conditioned on the current frame (per frame) as used in~\citet{shi2024amdm}. This is a special case of primitive with history length $H=1$ and future length $F=1$. The ablative model has significantly worse R-Prec and FID metrics and cannot respond to text prompts. This indicates using frame-by-frame prediction is less effective than primitives with a reasonable horizon ($H=2$, $F=8$) in learning text-conditioned motion space.

\methodname{} can learn high-quality text-conditioned motion primitive models using very few diffusion steps because of the simplicity of the primitive representation. We conduct ablative studies of training \methodname{} using different numbers of DDPM diffusion steps. 
Reducing the diffusion steps from 100 to fewer than 10 does not significantly affect performance. However, an extremely low diffusion step number of 2 leads to a much higher FID, indicating poorer motion quality.

\begin{table}[h]
\centering
\scriptsize
\setlength\tabcolsep{2.5pt}
\caption{Ablation studies results on text-conditioned temporal motion composition. 
The first row includes the metrics of the dataset for reference.
Symbol `$\rightarrow$' denotes that closer to the dataset reference is better and `$\pm$'  indicates the 95\% confidence interval.
}
\begin{tabular}{l|cccc|cccc}
\hline
 & \multicolumn{4}{c|}{Segment} & \multicolumn{4}{c}{Transition} \\
 & FID$\downarrow$ & R-prec$\uparrow$ & DIV $\rightarrow$ & MM-Dist$\downarrow$ & FID$\downarrow$ & DIV $\rightarrow$ & PJ$\rightarrow$ & AUJ $\downarrow$  \\
\hline
Dataset & 0.00$\pm$0.00 & 0.72$\pm$0.00 & 8.42$\pm$0.15 & 3.36$\pm$0.00 & 0.00$\pm$0.00 & 6.20$\pm$0.06 & 0.02$\pm$0.00 & 0.00$\pm$0.00\\
\hline
Ours & 3.79$\pm$0.06 & 0.62$\pm$0.01 & 8.05$\pm$0.10 & 5.27$\pm$0.01 & 1.86$\pm$0.05 & 6.70$\pm$0.03 & 0.06$\pm$0.00 & 0.21$\pm$0.00 \\
\methodname{}-VAE       & 4.23$\pm$0.02 & 0.62$\pm$0.01 & 8.33$\pm$0.12 & 5.29$\pm$0.01 & 1.79$\pm$0.02 & 6.73$\pm$0.23 & 0.20$\pm$0.00 & 0.96$\pm$0.00 \\
\methodname{}-schedule        & 8.08$\pm$0.09 & 0.39$\pm$0.01 & 8.05$\pm$0.12 & 6.96$\pm$0.03 & 7.41$\pm$0.10 & 6.58$\pm$0.06 & 0.03$\pm$0.00 & 0.18$\pm$0.00 \\
per frame($H$=1,$F$=1)  & 10.31$\pm$0.09 & 0.29$\pm$0.01 & 6.82$\pm$0.13 & 7.41$\pm$0.01 & 7.82$\pm$0.09 & 6.03$\pm$0.08 & 0.02$\pm$0.00 & 0.08$\pm$0.00 \\
\update{$H$=2,$F$=16}  & \update{4.04$\pm$0.10} & \update{0.66$\pm$0.00} & \update{8.20$\pm$0.06} & \update{4.96$\pm$0.01} & \update{2.22$\pm$0.10} & \update{6.60$\pm$0.20} & \update{0.06$\pm$0.00} & \update{0.18$\pm$0.00} \\
\hline
steps 2     & 4.44$\pm$0.04 & 0.60$\pm$0.00 & 8.20$\pm$0.15 & 5.38$\pm$0.01 & 2.24$\pm$0.02 & 6.77$\pm$0.07 & 0.05$\pm$0.00 & 0.20$\pm$0.00 \\
steps 5     & 3.49$\pm$0.09 & 0.63$\pm$0.00 & 8.25$\pm$0.15 & 5.18$\pm$0.01 & 2.11$\pm$0.07 & 6.74$\pm$0.11 & 0.05$\pm$0.00 & 0.20$\pm$0.00 \\
steps 8     & 3.70$\pm$0.03 & 0.62$\pm$0.01 & 8.04$\pm$0.13 & 5.25$\pm$0.03 & 2.15$\pm$0.08 & 6.72$\pm$0.15 & 0.06$\pm$0.00 & 0.20$\pm$0.00 \\
steps 10 (Ours)    & 3.79$\pm$0.06 & 0.62$\pm$0.01 & 8.05$\pm$0.10 & 5.27$\pm$0.01 & 1.86$\pm$0.05 & 6.70$\pm$0.03 & 0.06$\pm$0.00 & 0.21$\pm$0.00 \\
steps 50    & 3.82$\pm$0.05 & 0.60$\pm$0.00 & 7.74$\pm$0.07 & 5.30$\pm$0.01 & 2.11$\pm$0.10 & 6.58$\pm$0.10 & 0.06$\pm$0.00 & 0.22$\pm$0.00 \\
steps 100   & 4.16$\pm$0.06 & 0.61$\pm$0.00 & 7.82$\pm$0.15 & 5.32$\pm$0.02 & 2.20$\pm$0.05 & 6.43$\pm$0.10 & 0.06$\pm$0.00 & 0.21$\pm$0.00 \\
\hline
\end{tabular}

\label{tab:ablation}
\end{table}



\section{Latent diffusion noise optimization-based control}
\label{sec:app_optim}
\textbf{Optimization details.}
We use the Adam \citep{kingmaAdamMethodStochastic2017} optimizer with a learning rate of 0.05 to optimize latent noises $\mathbf{Z}_T$ for 100 to 300 steps. The learning rate is linearly annealed to 0 and the gradient is normalized to stabilize optimization.
\update{
The latent noises are initialized by sampling from Gaussian distributions. Empirically, we observe that initializing these latent noises with a small variance, such as 0.01, slightly improves the jerk and skate metrics at the cost of reduced motion diversity in the in-between experiment, compared to a unit variance initialization.
}
The optimization running time is dependent on both the motion sequence length and the number of optimization steps. An example experiment of a target motion sequence of 60 frames and 100 optimization steps costs around 74 seconds.

\textbf{Human-scene interaction.}
In human-scene interaction synthesis, we use two scene constraints to avoid human-scene collision and foot-floating artifacts.
To reduce human-scene interpenetration, $cons\_coll(\cdot)$ penalizes joints that collide into scenes and exhibit negative SDF values in each frame:
\begin{equation}
    cons\_coll(\mathbf{J}) = - \sum_{k=1}^{22}(\Psi(\mathbf{J}^k) - \boldsymbol{\tau}^k)_{-}
\end{equation}
where $\mathbf{J}^k$ denotes the location of the $k$-th joint in the scene coordinates, $\Psi(\cdot)$ denotes the signed distance function returning the signed distance from the query joint location to the scene, $\boldsymbol{\tau}$ denotes the joint-dependent contact threshold values, which are determined by the joint-skin distance in the rest pose, and $(\cdot)_{-}$ denotes clipping positive values.
To reduce the occurrence of foot-floating artifacts, $cons\_cont(\cdot)$ encourages the foot to be in contact with the scene in every frame and is defined as:
\begin{equation}
    cons\_cont(\mathbf{J}) = (\min_{k \in Foot} \Psi(\mathbf{J}^k) - \tau)_{+}
\end{equation}
where $Foot$ denotes the set of foot joint indices, $\tau$ denotes the foot-floor contact threshold distance, and $(\cdot)_{+}$ denotes clipping negative values. 

\section{Reinforcement Learning-based Control}
\label{sec:app_rl}
\textbf{Architecture.}
Both the actor and critic networks are 4-layer MLPs with residual connections and a hidden dimension of 512. 
We apply the tanh scaling: $x=4 \cdot tanh(x)$ \citep{lingCharacterControllersUsing2020d} to the actor output to clip the action prediction in the range of $[-4, 4]$, avoiding unbounded action predictions. 
The actor networks are initialized with close to zero weights to boost policy training following \citep{andrychowicz2020matters}.
The observation input contains a 512D text embedding, a 552D of history motion, a 1D observation of floor height relative to the human pelvis, a 1D observation of the floor plane distance from the pelvis to the goal location, and a 3D unit vector of the goal direction in the human-centric coordinates frames.
We clamp the goal distance observation with a maximum value of 5m and the goal direction in the range of a 120-degree field of view to simulate egocentric human perception.

\textbf{Rewards.}
The text-conditioned goal-reaching control policies are trained to maximize the discounted expectation of the following rewards \citep{zhang_wanderings_2022,li2024egogen,zhaoSynthesizingDiverseHuman2023a}:

We use three distance-related rewards to encourage the human agent to minimize its distance to the goal location.
\begin{equation}
    \update{r_{dist} = d^{i-1} - d^{i}}
\end{equation}
The distance reward $r_{dist}$ encourages the human to get closer to the goal location, where $d^i$ is the 2D distance between the human pelvis and the goal location at step $i$.

\begin{equation}
r_{succ} =
\begin{cases} 
1 & \text{if } d^i < 0.3 \\
0 & \text{otherwise}
\end{cases}
\end{equation}
The success reward $succ$ gives a sparse but strong reward when the human arrives at the goal, where 0.3m is the success threshold value.

\begin{equation}
r_{ori}=\frac{\langle \mathbf{p}^i - \mathbf{p}^{i-1}, g - \mathbf{p}^{i-1} \rangle + 1}{2},
\label{equ:ori_reward}
\end{equation}
The moving orientation reward encourages the moving orientation to be aligned with the goal orientation, where $\mathbf{p}^i$ is the human pelvis location at step $i$, and $g$ is the goal location.

Moreover, we apply scene constraints-related rewards to discourage unnatural behaviors such as foot skating and collision with the floor.
\begin{equation}
r_{skate}=-disp \cdot (2-2^{h/0.03}),
\label{equ:skate_reward}
\end{equation}
 The skate reward $r_{skate}$ penalizes foot displacements when in contact with the floor, where $disp$ is the foot displacement in two consecutive frames, $h$ denotes the higher foot height in two consecutive frames and 0.03m is the threshold value for contact.

\begin{equation}
r_{floor}=-(|lf| - 0.03)_{+},
\label{equ:floor_reward}
\end{equation}

The floor contact reward $r_{floor}$ penalizes when the lower foot distance to the floor is above the threshold of 0.03m, where $lf$ denotes the height of the lower foot, $|\cdot|$ denotes the absolute value operator, and $(\cdot)_{+}$ is the clipping operator with a minimum of 0. 

The rewards are weighted with $w_{dist}=1$, $w_{succ}=1$, $w_{ori}=0.1$, 
\update{$w_{skate}=100$, $w_{floor}=10$ for hopping and running and $w_{floor}=100$ for walking.}

\begin{figure}[ht]
\centering
\begin{minipage}{\textwidth}
    \centering
    \begin{subfigure}[b]{0.45\textwidth}
        \centering
        \includegraphics[width=\textwidth, trim=0cm 0cm 0cm 0.85cm, clip]{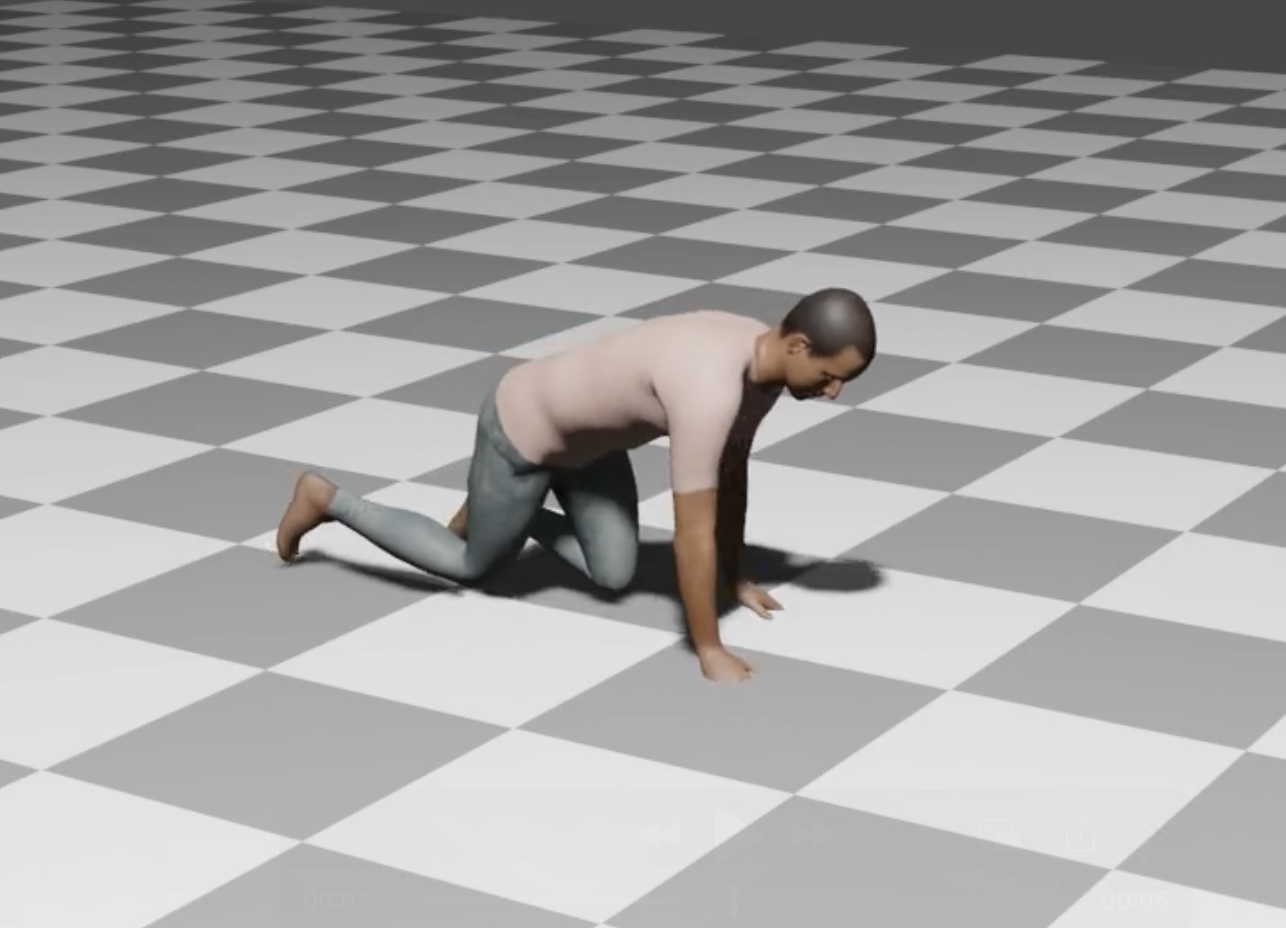}
        \caption{\update{Crawling sequence generated by \methodname{}}}
    \end{subfigure}
    \begin{subfigure}[b]{0.45\textwidth}
        \centering
        \includegraphics[width=\textwidth]{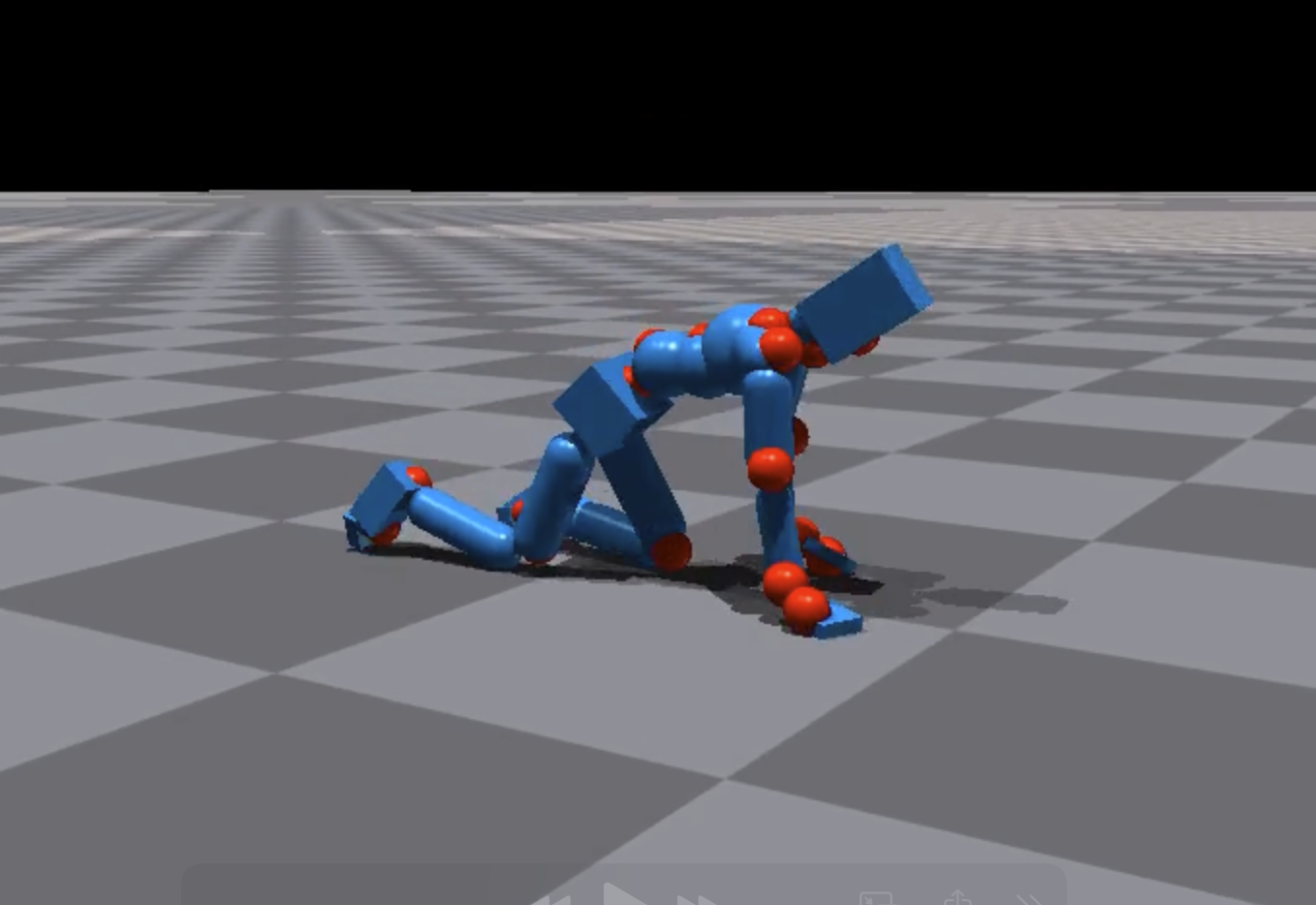}
        \caption{\update{Physics-based motion tracking result}}
    \end{subfigure}
    \caption{\update{We demonstrate an example of integrating \methodname{} with the physics-based motion tracking method PHC~\citep{Luo2023PerpetualHC} to achieve more physically plausible motions. The left image illustrates a crawling sequence generated by \methodname{}, exhibiting artifacts such as hand-floor penetration. The right image displays the physics-based motion tracking outcome applied to the raw generated sequence, which enhances joint-floor contact and resolves the hand-floor penetration issue.}}
    \label{fig:physics}
\end{minipage}
\end{figure}

\update{
\section{Combination with physics-based motion tracking}
}
\update{
Our method, \methodname{}, enables real-time motion generation in response to online text prompts. However, as a kinematic-based approach, \methodname{} may produce physically inaccurate motions with artifacts such as skating and floating. To address this, we demonstrate that \methodname{} can be integrated with physically simulated motion tracking methods, specifically PHC~\citep{Luo2023PerpetualHC}, to generate more physically plausible motions.
In Fig.~\ref{fig:physics}, we present an example sequence of a person crawling. The raw generation results from \methodname{} exhibit artifacts such as hand-floor penetration. Applying physics-based tracking to refine the raw motion successfully produces more physically plausible results, improving joint-floor contact and eliminating penetration artifacts. 
This integration combines the versatile text-driven motion generation of \methodname{} with the physical accuracy provided by the physics-based simulation.
Given the real-time capabilities of both \methodname{} and PHC, it is possible to leverage physics to correct the kinematic motion generated by \methodname{} on the fly and then use the corrected motions for subsequent online generation.
}

\update{
\section{Discussion of long rollout results given a single text prompt}
}
\update{
Our \methodname{} can autoregressively generate perpetual rollouts of actions that are inherently repeatable and extendable. For example, \methodname{} can produce minutes-long sequences of continuous human motion, such as jogging in circles, performing cartwheels, or dancing. 
These actions are inherently repeatable, and such extensions are also represented in the AMASS~\citep{AMASS:ICCV:2019} dataset.
\methodname{} can stably generate minutes-long rollouts of the same action, enabled by its autoregressive motion primitive modeling and scheduled training scheme.\\
Some other actions, however, have inherent boundary states that mark the completion of the action. For instance, ``kneel down'' reaches a boundary state where the knees achieve contact with the floor. Further extrapolation of ``kneel down'' beyond this boundary state is not represented in the dataset and is not intuitively anticipatable by humans, as no further motion logically extends within the action semantics. Continuing rollout using the ``kneel down'' text prompt results in motions exhibiting fluctuations around the boundary state.\\
In summary, long rollout results given a single text prompt will repeat naturally or fluctuate around a boundary state, depending on the inherent nature of the action and its representation in the dataset. 
We provide video results of minutes-long rollout generation results on our \href{https://zkf1997.github.io/DART/}{project website}.
}

\update{
\section{Discussion on open-vocabulary motion generation}
}

\update{
Limited vocabulary is a critical limitation and challenge shared by existing text-conditioned motion generation methods. Existing methods, including our approach DART, struggle to generalize to open-vocabulary text prompts due to the scarcity of 3D human motion data with text annotations. The scale of motion data available is several orders of magnitude smaller than that for text-conditioned image and video generation, primarily due to the reliance on marker-based motion capture systems, which are challenging to scale.\\
To expand the dataset and enable open-vocabulary generation, extracting human motion data from in-the-wild internet videos and generative image/video models~\citep{kapon2024mas, goel2023humans, lin2023motionx, shan2024towards}, is a promising direction. Additionally, the rapid advancement of vision-language models (VLMs) holds promise for automatically providing detailed, frame-aligned motion text labels to facilitate text-to-motion generation~\citep{shan2024towards, dai2023instructblip}.
}

\section{Computing resources}
\label{sec:computing}

Our experiments and performance profiling are conducted on a workstation with single RTX 4090 GPU, intel i7-13700K CPU, 64GiB memory. The workstation runs with Ubuntu 22.04.4 LTS system.
